\documentclass[10pt,journal,compsoc]{IEEEtran}

\ifCLASSOPTIONcompsoc
  \usepackage[nocompress]{cite}
\else
  \usepackage{cite}
\fi

\usepackage{multirow}
\usepackage{wrapfig}
\usepackage{graphicx}
\usepackage[table]{xcolor}
\definecolor{RevisedRows}{rgb}{1,0.88,1}
\usepackage{verbatim}
\usepackage{amsmath}
\usepackage{hyperref}
\usepackage[percent]{overpic}
\usepackage{pifont}
\usepackage{algorithm}
\usepackage{algpseudocode}
\usepackage{pdfpages}
\usepackage{multido}

\definecolor{module1}{rgb}{0.35, 0.61, 0.83}
\definecolor{module2}{rgb}{0.96,0.69,0.51}
\definecolor{module3}{rgb}{0.66,0.82,0.56}
\definecolor{module4}{rgb}{1,1,0.7}
\definecolor{module5}{rgb}{1,0.66,0.93}
\definecolor{module6}{rgb}{0.86,0.86,0.86}

\def\eg{\emph{e.g.}}

\def\ie{\emph{i.e.}}
\def\etal{\emph{et al.}}
\def\kitti{KITTI}
\def\about{$\sim$}
\newcommand{\xmark}{\ding{55}}%
\def\netname{\emph{MADNet}}
\def\extendednetname{\underline{M}odularly \underline{AD}aptive \underline{Net}work}
\def\algoname{\emph{MAD}}
\def\extendedalgoname{\underline{M}odular \underline{AD}aptation}
\def\kitti{KITTI}

\newcommand{\rev}[1]{ \color{black} #1 \color{black}}

\begin{document}

\title{Continual Adaptation for Deep Stereo}

\author{Matteo~Poggi,~\IEEEmembership{Member,~IEEE,}
        Alessio~Tonioni,~\IEEEmembership{Member,~IEEE,}
        Fabio~Tosi,~\IEEEmembership{Student Member,~IEEE,}
        Stefano~Mattoccia,~\IEEEmembership{Member,~IEEE,}
        and~Luigi~Di Stefano,~\IEEEmembership{Member,~IEEE}%
\IEEEcompsocitemizethanks{\IEEEcompsocthanksitem M. Poggi, F. Tosi, S. Mattoccia and L. Di Stefano are with the Department
Computer Science and Engineering, University of Bologna, Italy.%
\IEEEcompsocthanksitem A. Tonioni is with Google Zurich.}%
}

\IEEEtitleabstractindextext{%
\begin{abstract}
Depth estimation from stereo images is carried out with unmatched results by convolutional neural networks trained end-to-end to regress dense disparities. Like for most tasks, this is possible if large amounts of labelled samples are available for training, possibly covering the whole data distribution encountered at deployment time.
Being such an assumption systematically unmet in real applications, the capacity of \textit{adapting} to any unseen setting becomes of paramount importance.
Purposely, we propose a continual adaptation paradigm for deep stereo networks designed to deal with challenging and ever-changing environments. We design a lightweight and modular architecture, \extendednetname{} (\netname{}), and  formulate \extendedalgoname{} algorithms (\algoname{}, \algoname{}++)  which permit efficient optimization of independent sub-portions of the entire network. In our paradigm, the learning signals needed to continuously adapt  models online can be sourced from self-supervision via right-to-left image warping or from traditional stereo algorithms. With both sources, no other data than the input images being gathered at deployment time are needed.
Thus, our network architecture and adaptation algorithms realize  the first real-time  self-adaptive deep stereo system and pave the way for a new paradigm that can facilitate practical  deployment of end-to-end architectures for dense disparity regression.
\end{abstract}

}

\maketitle

\IEEEdisplaynontitleabstractindextext

\IEEEpeerreviewmaketitle

\IEEEraisesectionheading{\section{Introduction}\label{sec:introduction}}

\IEEEPARstart{E}{}stimating dense and accurate depth maps is a key perception step to pursue scene comprehension tasks dealing with navigation and interaction with the environment. Passive, image-based techniques aimed at depth perception compare favourably to active sensors in terms of cost, bulkiness as well as - more often than not- working range and flexibility. Among such techniques, stereo vision  \cite{scharstein2002taxonomy} is usually the preferred choice, requiring just a pair of synchronized and calibrated cameras to measure depth by triangulation between matching pixels.

Akin to most computer vision problems, in the last years deep learning has entered into solutions for stereo matching, at first replacing certain specific steps of the pipeline by neural networks (\eg, matching cost computation \cite{zbontar2016stereo}) then rapidly converging toward end-to-end architectures \cite{mayer2016large,Kendall_2017_ICCV}. Although end-to-end networks have established the new state-of-the-art in challenging benchmarks such as \kitti{} \cite{KITTI_2012,KITTI_2015}, they require a large amount of images labelled with ground-truth disparities, \rev{in order to effectively learn how to tackle stereo.} 
As obtaining ground-truth disparities, \ie{} depths, for real images is particularly challenging and expensive, computer graphics has became a popular alternative to gather thousands of \textit{synthetic} images endowed with depth labels for free  \cite{mayer2016large}. Although highly realistic, these images can hardly encompass all the nuisances occurring in the real world, such as, e.g., sensor noise, reflective surfaces and challenging illumination conditions. Thus, due to the \textit{domain shift} between the training and testing environments \cite{Tonioni_2017_ICCV}, deep networks trained by computer-generated imagery suffer from a large loss in accuracy when deployed in the real world. A partial solution to this issue consists in fine-tuning the stereo network on few labelled samples from the real domain. Yet, to obtain such ground-truth labels, costly active sensors (\eg, LIDAR) and manual intervention or post-processing  are required \cite{Uhrig2017THREEDV}. Even more importantly, despite fine-tuning by a few real-images may address  the \textit{synthetic-to-real} domain shift, it cannot take into account the countless diverse environmental conditions that a stereo network meant to be deployed \textit{in-the-wild} may encounter, such as, in autonomous driving scenarios, urban and countryside roads, tunnels, varying weather and sudden changes of the surroundings.  

\begin{figure}[t]
    \centering
    \renewcommand{\tabcolsep}{1pt}
    \begin{tabular}{cc}
        \begin{overpic}[width=0.22\textwidth]{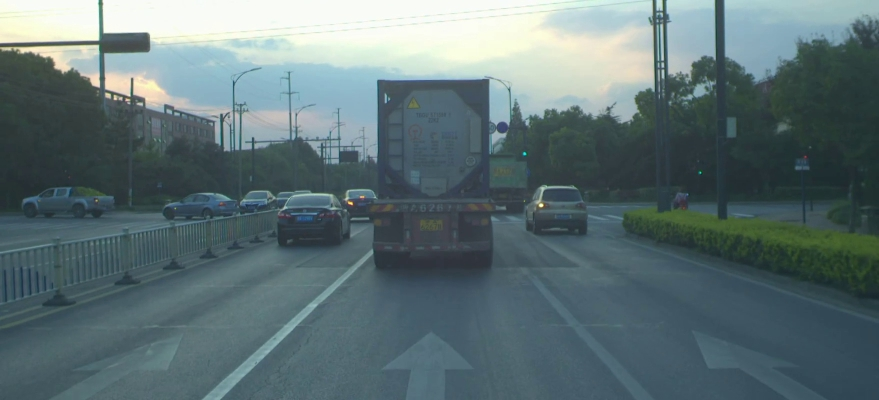} 
        \put (88,38) {$\displaystyle\textcolor{white}{\textbf{(a)}}$}
        \end{overpic} &  
        \begin{overpic}[width=0.22\textwidth]{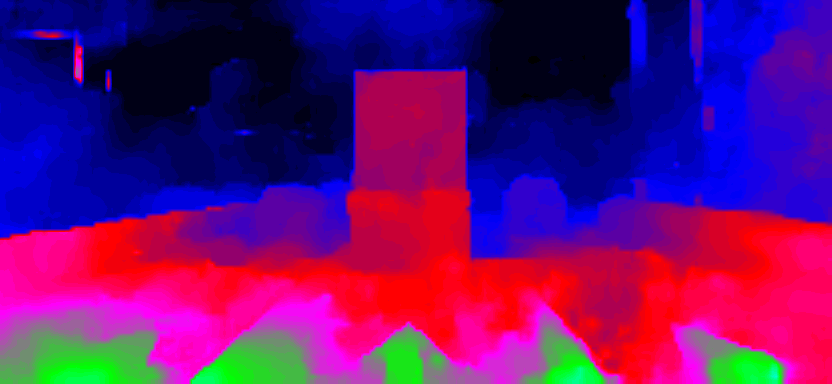}\put (88,38) {$\displaystyle\textcolor{white}{\textbf{(b)}}$}
        \end{overpic} \\
        \begin{overpic}[width=0.22\textwidth]{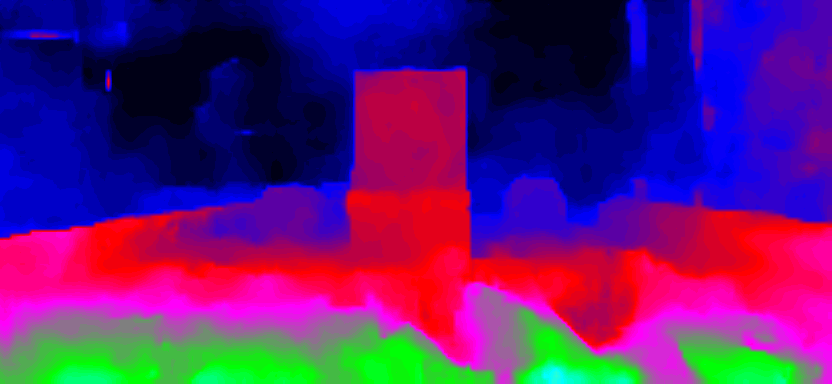}
        \put (88,38) {$\displaystyle\textcolor{white}{\textbf{(c)}}$}
        \end{overpic} &
        \begin{overpic}[width=0.22\textwidth]{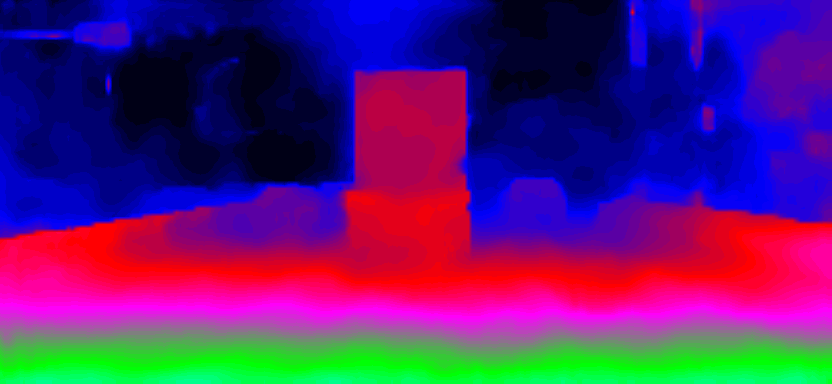}
        \put (88,38) {$\displaystyle\textcolor{white}{\textbf{(d)}}$}
        \end{overpic} \\
    \end{tabular}
    \caption{\textbf{Continual adaptation on real images.} We show the reference image of a stereo pair from DrivingStereo  \cite{Yang_2019_CVPR} (a) and the disparity maps computed by \netname{} when trained on synthetic data only (b) or adapted online by either \algoname{} (c)  or \algoname{}++ (d). }
    \label{fig:intro}
\end{figure}

In our vision, the only viable approach to practical end-to-end deep stereo deals with departing from the traditional \textit{training-validation-testing} workflow towards a \textit{continual adaptation} paradigm, so as to realize neural networks able to \textit{adapt} promptly to new situations and environments. Of course, this novel paradigm cannot leverage standard supervised learning approaches since ground-truth labels would not be available for any new setting faced \emph{in-the-wild}.

In this paper we extend our preliminary work on \textit{continual adaptation} for deep stereo \cite{Tonioni_2019_CVPR}, which proposed the first-ever real-time, self-adapting, deep stereo network by relying on self-supervision obtained from the input pair of frames via a photometric loss \cite{zhou2017unsupervisedStereo,godard2017unsupervised,zhang2018activestereonet}. 
Given a sequence of stereo pairs, a straightforward continual adaptation scheme might be realized though the following steps:  1) output the prediction and compute the loss for the current pair of frames; 2) update the whole network by back-propagation; 3) move forward to the next pair of frames, with enriched knowledge encoded in the updated weights. 
However, due the computational overhead associated with on-line back-propagation, most state-of-the-art stereo architectures would not operate in real-time under a \textit{continual adaptation} paradigm mandating full update of the network via back-propagation. Thus, we designed a \extendednetname{} (\netname{}) architecture that is fast, accurate and features small memory requirements. Moreover, we developed a \extendedalgoname{} (\algoname{}) algorithm that, in each on-line back-propagation step, enables to select and update only a portion of the whole \netname{}, thereby vastly reducing the overhead required by online model updating and permitting prediction alongside self-adaptation in real-time without a large loss in accuracy compared to back-propagating errors into the full set of weights of network.  

We extend and improve the  \netname{}-\algoname{} framework by leveraging on \textit{proxy supervision} obtained from traditional stereo algorithms  \cite{Tonioni_2017_ICCV,tonioni2019unsupervised}. Indeed, although deep stereo networks are unmatched when trained and tested on similar domains, traditional methods, like \cite{hirschmuller2005accurate}, are largely \textit{domain-agnostic}. In fact, they behave similarly and suffer from the same nuisances (\eg{}, low-textured regions, occlusions, repetitive patterns) with both synthetic and real images as well as across diverse environmental settings. 
This suggests that traditional algorithms may be exploited to supervise online deep stereo networks if matching errors, \ie{} outliers, are properly detected and filtered out. We show that this strategy yields  a much stronger adaptation ability and that this results consistently in a significant shrink of the performance gap between modular and full adaptation  of \netname{}, with the former often turning out even more beneficial than the latter.  
As a matter of fact, \autoref{fig:intro} shows a qualitative comparison between the \algoname{} formulation described in \cite{Tonioni_2019_CVPR} and the novel approach proposed in this paper, referred to hereinafter as  \algoname{}++.
Moreover, we dig into our \textit{continual adaptation} paradigm and thoroughly explore its behaviour across very different datasets, showing, in particular, that our proposal  is not affected by \textit{catastrophic forgetting} while, on the contrary, continually adapting the stereo network is beneficial to performance in case of domain changes.  

The main contributions of this unabridged paper on our work on continual adaptation for deep stereo can be summarized  as follows.
\begin{itemize}
    \item We introduce \algoname{}++  which ameliorates our \algoname{} framework by leveraging on proxy supervision provided by traditional stereo algorithms. This novel approach outperforms the original proposal and yields often better results than updating all network weights while running twice faster.
    
    \item We extensively evaluate  both \algoname{} and \algoname{}++ on the raw \kitti{} dataset already considered in  \cite{Tonioni_2019_CVPR}. Besides, we include experiments on two additional datasets, \ie{} DrivingStereo \cite{Yang_2019_CVPR} and WeanHall \cite{weanhall}, so as to provide stronger evidence on the effectiveness of our proposed framework in a broader variety of target domains. 
    
    \item We test our methods across datasets to highlight how continual learning is robust to domain shifts. We find  no evidence of catastrophic forgetting in any experiments. On the contrary, we show that models adapted elsewhere feature better adaptation ability when facing new domains.
    
\end{itemize}

\section{Related work}\label{sec:related}

In this section, we briefly review the literature relevant to our work.

\textbf{Machine learning for stereo.} The first attempts to use machine learning for stereo matching dealt with estimating confidence measures \cite{Poggi_2017_ICCV}, either by random forest \cite{Hausler_2013_CVPR,Spyropoulos_2014_CVPR,Park_2015_CVPR,Poggi_2016_3DV} or  CNNs \cite{Poggi_2016_BMVC,Seki_2016_BMVC,TOSI_2018_ECCV,Kim_2019_CVPR}, and often with the purpose of improving the final accuracy when combined with traditional algorithms. Regarding stereo algorithms, the first works proposed matching cost functions realized by patch-based CNNs \cite{zbontar2016stereo,Chen_2015_ICCV,luo2016efficient} and allowed to achieve state-of-the-art performance by replacing handcrafted  cost functions \cite{SCHARSTEIN_COST} within the SGM pipeline \cite{hirschmuller2005accurate}. Later, Batsos \etal \cite{batsos2018cbmv} combined traditional matching functions within a random forest framework to obtain  better generalization across domains.
Then, Shaked and Wolf \cite{Shaked_2017_CVPR} proposed to rely on deep learning for  matching cost computation, disparity selection and confidence prediction, while Gidaris and Komodakis \cite{Gidaris_2017_CVPR} focused on disparity refinement.

A true paradigm shift did occur with the first end-to-end stereo network, DispNetC, was proposed alongside large synthetic training datasets \cite{mayer2016large}. In \cite{mayer2016large} a custom correlation layer was  designed to encode the similarities between pixels as features. Kendall \etal \cite{Kendall_2017_ICCV} designed GC-Net, switching towards 3D networks that build a cost volume by means of features concatenation.
These two architectures started the development of two main families of networks, referred to as 2D and 3D, respectively.
Proposals belonging to the former class use typically  a single or multiple correlation layers \cite{Pang_2017_ICCV_Workshops,liang2018learning,Ilg_2018_ECCV,song2018edgestereo,yang2018segstereo,Tonioni_2019_CVPR,yin2019hierarchical,Duggal2019ICCV}, while 3D networks build 4D volumes by means of concatenation \cite{chang2018pyramid,Nie_2019_CVPR,tulyakov2018practical,wang2019anytime,Zhang2019GANet,zhang2020adaptive,cheng2019learning,cheng2020hierarchical,xu2020aanet}, features difference \cite{khamis2018stereonet} or group-wise correlations \cite{guo2019group}, both combined with active sensors such as LIDAR in \cite{Poggi_CVPR_2019}.
Although most works focus on accuracy, others deploy lightweight architectures \cite{Tonioni_2019_CVPR,wang2019anytime,khamis2018stereonet} aimed at real-time performance, sometimes combining stereo with semantic segmentation \cite{Dovesi_ICRA_2020} or pursuing scene flow \cite{Jiang_2019_ICCV,aleotti2020learning}.
Unfortunately, however, all end-to-end stereo networks are prone to domain shift, as performance decay dramatically when the model is run in environments different from those observed at training time, as shown in \cite{Tonioni_2017_ICCV,tonioni2019unsupervised,pang2018zoom,Tonioni_2019_CVPR,Tonioni_2019_learn2adapt}.

\textbf{Self-supervision from photometric losses}. View synthesis has been recently used to train depth estimation networks in a self-supervised manner by photometric losses \cite{godard2017unsupervised,zhou2017unsupervised}. For monocular depth estimation, multiple images are deployed at training time in order to replace ground-truth labels by warping the different views, coming either from stereo pairs or image sequences, according to the predicted depth and minimizing the photometric error between real and warped images \cite{garg2016unsupervised,zhou2017unsupervised,godard2017unsupervised,pydnet18,3net18}.
Other recent works follow a similar approach for deep stereo matching \cite{zhou2017unsupervisedStereo,zhang2018activestereonet,zhong2018open,wang2019unos,lai19cvpr,Tonioni_2019_CVPR}. Unlike monocular ones, however, in  stereo setups the input images used to compute the photometric loss are available at both training and testing time, which renders this self-supervised learning protocol amenable to \textit{continual adaptation}. 

\textbf{Proxy-supervision from distilled labels}. A further approach consists in sourcing  \textit{pseudo} ground-truth annotations, namely \textit{proxy labels}, accurate enough to allow for effective supervision during training. The process to obtain these annotations is usually referred to as \textit{distillation}. In the field of depth estimation, the work by Tonioni \etal \cite{Tonioni_2017_ICCV,tonioni2019unsupervised} was the first to use traditional stereo algorithms filtered out by means of confidence measures for offline adaptation of deep stereo networks. Pang \etal \cite{pang2018zoom} used iterative optimization over proxy labels of the network itself obtained at higher resolutions. Recently, these approaches have been applied to monocular depth estimation, sourcing proxies either from traditional stereo algorithms \cite{Tosi_2019_CVPR}, a teacher architecture \cite{Pilzer_2019_CVPR} or the network itself trained in two stages \cite{Poggi_2020_CVPR}.

\section{Continual Adaptation}
\label{sec:continual}

State-of-the-art deep stereo networks are severely challenged by \textit{Out-of-Distribution} generalization, frequently exhibiting large accuracy drops when deployed across different environments. This issue is typically alleviated by fine-tuning the network on additional labelled samples  from the target  distribution. We argue that this is definitely unpractical as 1) it requires collecting data and tuning the model before real deployment for any target environment and 2) ground-truth labels need to be acquired together with images.

\begin{algorithm}[t]
	\caption{Full Adaptation (\emph{FULL})}
	\label{algo:backprop}
	\begin{algorithmic}[1] 
		\State \textbf{Require:} Stereo model $\mathcal{N}$ parametrized by $\Theta$ 
		\State $t=0$
		\While {$not$ $stop$}
		\State $ x_{t} \gets ReadFrames(t) $
		\State $ y_{t} \gets ForwardPass(\mathcal{N}, \Theta_{t}, x_{t}) $ 
		\State $ \mathcal{L}_{t} \gets Loss(x_{t},y_{t}) $
		\State $ \Theta_{t+1} \gets UpdateWeights(\mathcal{L}_t,\Theta_t) $
		\State $t \gets t+1$
		\EndWhile	 
	\end{algorithmic}
\end{algorithm}

\textit{Self-supervision} and \textit{distillation} allow to circumvent the need for ground-truth labels, making offline fine-tuning the main obstacle towards seamless deployment in-the-wild. This can be addressed  by moving from a traditional \textit{train-validation-test} procedure to a \textit{continual adaptation} paradigm, whereby the distinction between offline training and online testing is relaxed due to both being performed online and at once.
In Algo \autoref{algo:backprop} we provide a description of a continual adaptation process referred to as \emph{Full Adaptation}. Given a stereo network $\mathcal{N}$ parametrized by a set of weights $\Theta$, at any given time frame $t$ we read a new stereo pair $x_t$, made out of a left and right image $(l_t,r_t)$, and predict a disparity map $y_t$ \rev{for image $l_t$} based on the current set of parameters $\theta_t$. Then, a suitable loss function $\mathcal{L}_{t}$ is computed from $x_t, y_t$ and used to update the network weights $\Theta$ before reading the next stereo pair $x_{t+1}$.

Since a train iteration is performed on-the-fly on each incoming stereo pair, the network always learns and - potentially- improves by gathering knowledge from the sensed environment. \rev{It is worth pointing out that this approach is different from standard practice dealing with updating a network on a batch of images to optimize it over a set of variegate samples that better approximate the entire training set and lead to more stable gradients. In our setting, however, the update steps occur on images that are acquired closely in time and are, therefore, very similar. Thus, the gradients provided by the individual samples are quite similar to the average gradient that would be computed in a batch. In the supplementary material we show that updating the network on a batch of samples does not provide significant benefits.}

As shown in \cite{Tonioni_2019_CVPR}, our straightforward formulation, though intuitive and effective, introduces a non-negligible computational overhead that does increase the network latency dramatically. To address this drawback, we introduce a modular neural network architecture and a learning algorithm, which are designed to work in synergy to achieve effective continual adaptation with a limited computational overhead.

As pointed out, updating the whole network to achieve continual adaptation is time consuming and may hinder applicability to real-word applications calling for  tight low-latency requirements. 
Due to the time required by back-propagation being proportional to the number of network layers to be traversed, we may speed-up the computation by having fewer layers, \ie{} fewer weights, to update. Intuitively this is similar to accelerating forward inference by early-stopping the network processing in order to calculate only a subset of the total number of operations \cite{pydnet18,wang2019anytime}. 
Our work leverages on a similar intuition to speed up online back-propagation,  \ie{} the main computational overhead introduced by continual adaptation. 

\begin{figure}[t]
    \centering
    \includegraphics[width=0.3\textwidth]{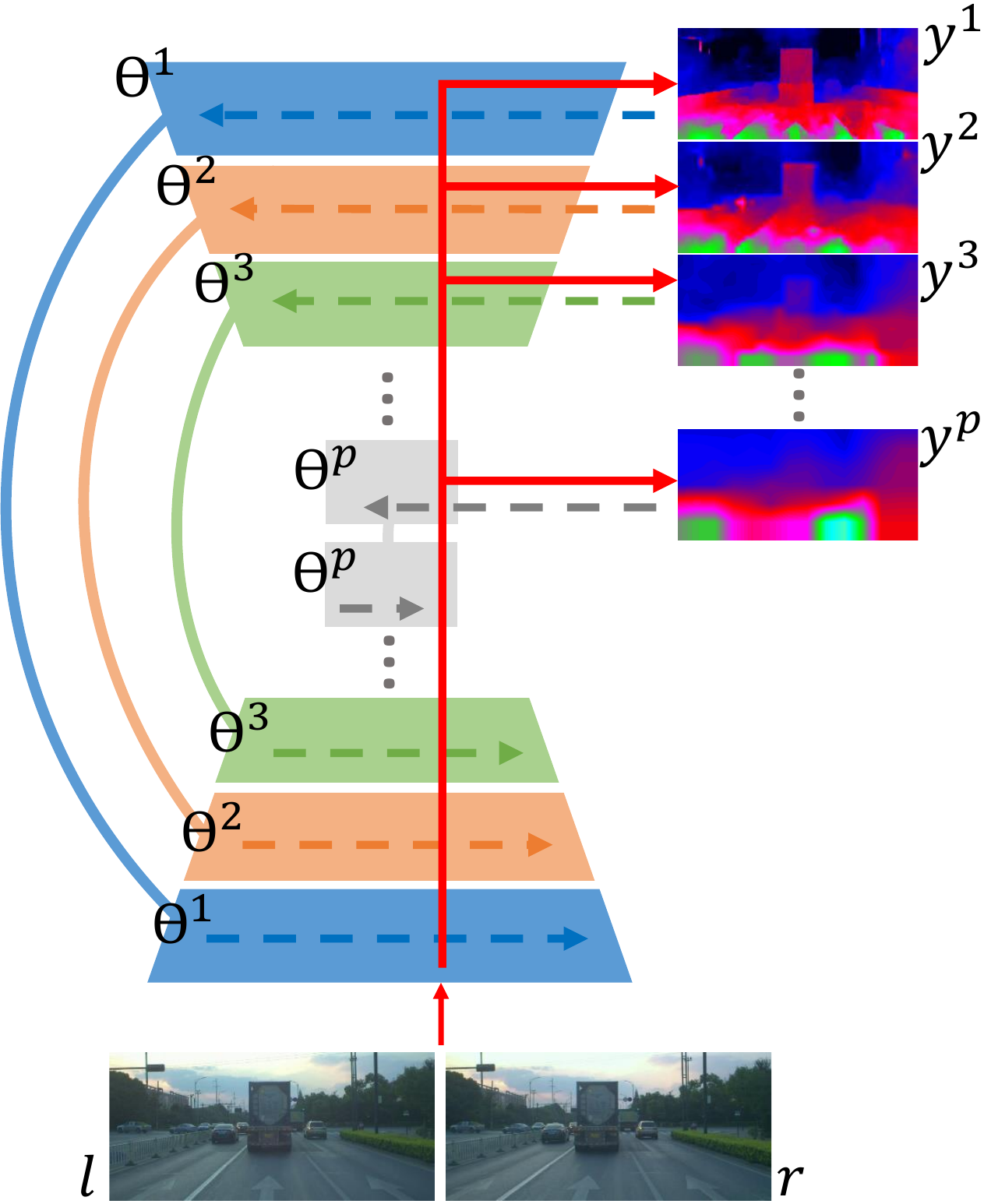}
    \caption{\textbf{Generic design of a modular adaptive network.} The network $\mathcal{N}$ is organized as a set of  non-overlapping modules $[\Theta^1,\dots,\Theta^p]$ and is trained to estimate a set of corresponding outputs $[y^1,\dots,y^p]$. During adaptation, a full forward pass (red line) is performed to obtain the outputs, on which losses $[\mathcal{L}^1,\dots,\mathcal{L}^p]$ are computed. By selecting a single $\mathcal{L}^i$, only one of the back-propagation routes (dashed lines) is followed so to update a single  module $\Theta^i$.}
    \label{fig:architecture}
\end{figure}

We start from the general encoder-decoder architecture  illustrated  in \autoref{fig:architecture}, \rev{which can be thought of as an abstraction of the majority of deep networks proposed for stereo matching in the past years.} 
The layers in network $\mathcal{N}$  can be organized into non-overlapping, but inter-connected, modules $[\Theta^1,\dots,\Theta^p]$ according to any arbitrary grouping policy, \eg, the resolution at which they operate, as shown in different colors in the figure. 
Then, we might think of performing back-propagation on only one module at a time to speed it up. However, standard network architectures provide only a prediction amenable to computing the loss  at the very last layer. Therefore to back-propagate the training signal to each module we would still need to go through the deeper ones in the architecture.  
To overcome this limitation, we introduce shortcut back-propagation routes for each module.
Our network predicts as many outputs as the modules, $[y^1,\dots,y^p]$, and includes at least one back-propagation route from each output to all the layers belonging to the associated module, \ie  $(y^i,\Theta^i)$ with $i \in [1,\dots,p]$. Hence, by computing a loss $\mathcal{L}^i$ for each output, $y^i$  we can directly back-propagate into the corresponding module $\Theta^i$ avoiding the remaining ones by means, for instance, of skip connections (depicted as arcs in \autoref{fig:architecture}). The gradients computed with this strategy are an approximation of the trues ones, but they do  provide a good training signal, as it will be shown experimentally. 

An example of such a design is detailed in \autoref{sec:madnet} and used in the experiments reported in this paper. In this architecture each $y^i$ is a disparity prediction at a different spatial resolution, with $y^1$ denoting the highest resolution disparity map delivered as output, while each $\Theta^i$ includes all the network layers processing features at that resolution, \ie, both in the encoder and the decoder. Due to all the layers in $\Theta^i$ being connected through at least a direct back-propagation path, we can approximate the gradients for all layers by back-propagating only through the direct connection and skipping all the other back-propagation routes. 
This paradigm \textit{approximates}  back-propagation into the whole network by updating layers through time, \eg{} in $p$ steps should the modules  be sequentially updated, while providing a fast inference time, as required by many practical applications. 

\subsection{\extendedalgoname{} -- \algoname{}}
\label{sec:mad}

To pursue the modular adaptation approach described in the previous section as effectively as possible, we have developed a selection strategy aimed at choosing the module  $\Theta^i$ to be updated  at each time step. Purposely, we have devised the reward/punishment algorithm outlined in Algo \autoref{algo:mad}. At bootstrap (2), a histogram $\mathcal{H}$ consisting of $p$ bins (one per module) is initialized to zero. Then, at each time step $t$ the disparity maps $[y^1,\dots,y^p]_t$ are predicted (6) and the corresponding losses $[\mathcal{L}^1,\dots,\mathcal{L}^p]_t$ computed (7). Then, we select a network module $\Theta^{\phi_t}$ by sampling an index $\phi_t$ from the probability distribution associated with $\mathcal{H}$ (8) and perform back-propagation into $\Theta^{\phi_t}$ only (9). At this point, our network has been updated and it is ready to process the next stereo pair. Before moving on, we also update  $\mathcal{H}$ in order to reward or punish the module updated in the previous time step, namely $\Theta^{\phi_{t-1}}$, depending on whether this has proven to be effective or not. \rev{To do so, we linearly extrapolate the expected value for the highest resolution loss (i.e., that computed on the final output of the network), $\Tilde{\mathcal{L}}^{1}$, from the previous ones at times $(t-1)$ and $(t-2)$ (13).} Then, we compute the difference, $\gamma$, between the expected and computed losses (14). In case of a positive/negative difference we deem the update step on the module selected at time $(t-1)$ to have been effective/ineffective as at time $t$ the highest resolution loss turns out smaller/larger than the value we would have expected had module $\Theta^{\phi_{t-1}}$ not been updated. Accordingly, \rev{we gradually apply a decay factor $\delta$ to all entries in $\mathcal{H}$ (15)}, then we reward/punish module $\phi_{t-1}$ by adding a contribution $\gamma$ into histogram bin $\mathcal{H}[\phi_{t-1}]$ (16).
\rev{In our experiments, we set $\delta$ and $\lambda$ to 0.99 and 0.01, respectively, and in the supplementary material we show how the sensitivity of the algorithm to these parameters is moderate.}

\begin{algorithm}[t]
	\caption{\extendedalgoname{} (\algoname{}, \algoname{}++)}
	\label{algo:mad}
	\begin{algorithmic}[1] 
		\State \textbf{Require:} Stereo model $\mathcal{N}$ parametrized by $[\Theta^1,\dots,\Theta^p]$  
		\State $\mathcal{H}=[h^1,\dots, h^p] \gets 0$ 
		\State $t=0$
		\While {$not$ $stop$}
		\State $ x_{t} \gets ReadFrames(t) $
		\State $ [y^1,\dots,y^p]_t \gets ForwardPass(\mathcal{N},\Theta_{t},x_{t}) $ 
		\State $ [\mathcal{L}^{1},\dots,\mathcal{L}^{p}]_t \gets Loss(x_t,[y^1,\dots,y^p]_t) $
		\State $ \phi_t \gets Sample(softmax(\mathcal{H})) $

		\State $ \Theta^{\phi_t}_{t+1} \gets UpdateWeights(\mathcal{L}^{\phi}_t,\Theta^{\phi_t}_t) $
		\If {$t==0$}
		\State $ \mathcal{L}^{1}_{t-2} \gets \mathcal{L}^{1}_{t}$,

		$ \mathcal{L}^{1}_{t-1} \gets \mathcal{L}^{1}_{t}$
		\EndIf
		\State $ \Tilde{\mathcal{L}}^{1}_{t} \gets 2 \cdot \mathcal{L}^{1}_{t-1} - \mathcal{L}^{1}_{t-2} $
		\State $ \gamma \gets \Tilde{\mathcal{L}}^{1}_{t} - \mathcal{L}^{1}_{t} $
		\State $ \mathcal{H} \gets \delta \cdot \mathcal{H} $
        \State $ \mathcal{H}[\phi_{t-1}] \gets \mathcal{H}[\phi_{t-1}] + \lambda \cdot \gamma $
		\State $ \mathcal{L}^{1}_{t-2} \gets \mathcal{L}^{1}_{t-1}$,
		$ \mathcal{L}^{1}_{t-1} \gets \mathcal{L}^{1}_{t}$,
		$ \phi_{t-1} \gets \phi_{t} $
		\State $t \gets t+1$
		\EndWhile	 
	\end{algorithmic}
\end{algorithm}

In \cite{Tonioni_2019_CVPR} we realized Algo \autoref{algo:mad} using as  $\mathcal{L}^i, i \in [1,\dots,p]$, the self-supervised loss provided by the photometric error between the left image $l_t$ and the right image $\Tilde{r}_t$ warped according to the estimated disparity $y^i_t$. In particular, according to a popular choice in literature, we compute this photometric error as

\begin{equation}
\label{eq:photo}
    \mathcal{L}^i_t = \alpha \cdot \frac{1-\text{SSIM}(l_t,\Tilde{r}_t)}{2} + (1-\alpha)|l_t - \Tilde{r}_t|
\end{equation}
with $\alpha$ set to 0.85 \cite{godard2017unsupervised}. Thus the approach refereed to as \algoname{}) in this paper performs real-time continual adaptation by deploying a popular self-supervised loss within Algo \autoref{algo:mad}.  Although fast and effective, \algoname{} consists in diluting over time the network optimization process, thereby requiring more frames (\ie, update  steps) than the straightforward full adaptation approach (Algo \autoref{algo:backprop}) to acquire the knowledge needed to adapt a model to a novel environment. In the next section we describe how to leverage on a different kind of loss which deploys a stronger source of supervision while still being amenable to continual adaptation. In \autoref{sec:expres} we will show how this novel formulation can effectively accelerate a network optimization process distributed over time according to Algo \autoref{algo:mad} and reduce the performance gap with respect to Algo \autoref{algo:backprop} dramatically.

\subsection{Proxy-Supervised \extendedalgoname{} -- \algoname{}++}
\label{sec:mad++}

To speed-up the model adaptation process, we move toward a stronger form of supervision. 
In particular, we propose to rely on \textit{proxy supervision} by leveraging on a reliable external source of disparities used as proxies for ground-truth labels. For instance, the use of active sensors, like LIDARs, have been proposed  to supervise a depth prediction network \cite{Kuznietsov_CVPR_2017}.  Yet, a cheaper - and far more practical- source of proxy labels is described in recent works concerning both stereo \cite{Tonioni_2017_ICCV,tonioni2019unsupervised} and monocular \cite{Tosi_2019_CVPR} depth estimation. Accordingly, the noisy disparities computed by traditional stereo algorithms are filtered by a confidence estimator and deployed as proxy ground-truth labels to either adapt or train from scratch a depth prediction model.
Since the procedure described in Algo \autoref{algo:mad} is agnostic to the actual loss function, we can extend our modular adaptation approach so as to rely on proxy supervision by simply specifying a suitable loss $\mathcal{L}^i, i \in [1,\dots,p]$. This  novel formulation of Algo \autoref{algo:mad} will be referred to hereinafter as \algoname{}++. 

Given a generic stereo matching pipeline $\mathcal{M}$, we can obtain a noisy disparity map $z_t$ by processing an input stereo pair $x_t=(l_t,r_t)$. However, as discussed in  \cite{Tonioni_2017_ICCV,tonioni2019unsupervised},  to effectively supervise a depth estimation network it is crucial to filter out most of the noisy disparities. This can be achieved by estimating a confidence map, $c_t$, encoding the reliability of each pixel in $z_t$ \cite{Tonioni_2017_ICCV,tonioni2019unsupervised}. Then, supervision for any estimated $y^i_t$ can be obtained from $z_t$ by the following loss function:

\begin{equation}
    \mathcal{L}^i_t = \eta_t \cdot | y^i_t - z_t |
\end{equation}

where $\eta_t$ denotes an indicator function that selects the measurements in $z_t$ characterized by a sufficiently high confidence, \eg  a threshold operator applied to each pixel $p$ according to the estimated confidence $c_t(p)$:

\begin{equation}\label{eq:eta}
    \eta_t(p)=
\begin{cases}
1 & \text{if } c_t(p) \geq \varepsilon \\
0 & \text{otherwise } \\
\end{cases}
\end{equation}

\begin{figure}
    \centering
    \includegraphics[width=0.3\textwidth]{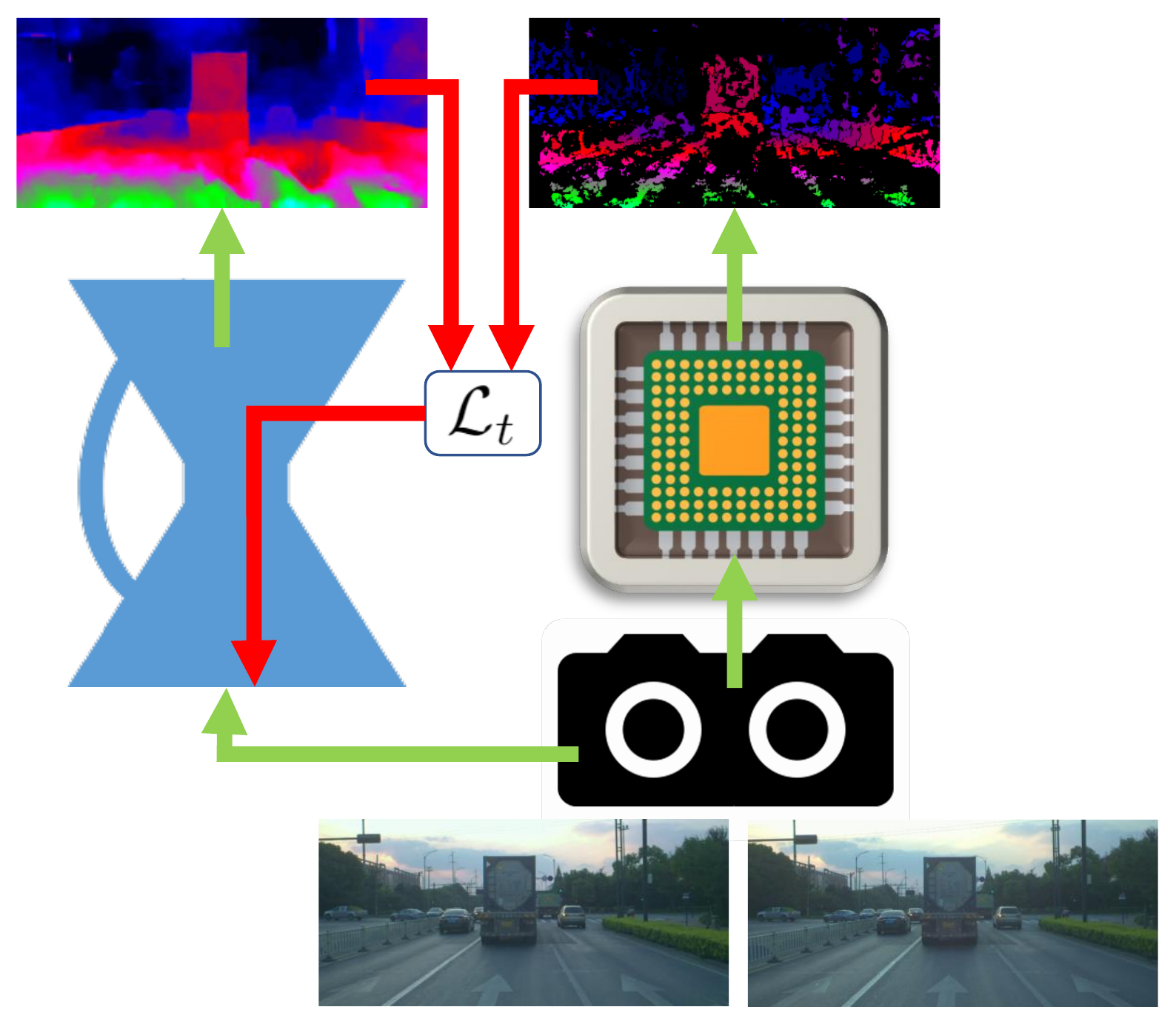}
    \caption{\textbf{Deployment of on-camera disparity computation within \algoname{}++.} During the forward pass (green arrows) the acquired frames are processed by \netname{}  to predict a disparity map as well as, in parallel, by a dedicated platform on-board the camera (\eg, an FPGA) to compute proxy disparity labels. During the backward pass (red arrows), the network is updated so as to minimize the loss given by the discrepancy between the predicted and proxy disparities.}
    \label{fig:rtsa++}
\end{figure}

\textbf{Stereo Matching.} Proxy labels can be obtained from an external stereo algorithm with a negligible overhead compared to the computational complexity of a deep stereo network. Indeed, a number of stereo cameras endowed with on-board processing hardware designed to deliver disparity maps at 50+ FPS  are available nowadays  \cite{HW_SGM_BANZ,HW_SGM_MERCEDES,Oberpfaffenhofen,smartcamera,rahnama2019real,rahnama2018r3sgm,POLLEFEYS_SGM}. As these cameras do not offload the stereo matching computation to the host device, they are amenable to distilling knowledge, \ie{} proxy ground-truth disparities, to a deep stereo network so as to run continual adaptation without slowing down the process.
\autoref{fig:rtsa++} illustrates how a stereo camera equipped with on-board processing can be deployed to support our real-time continual adaptation framework for deep stereo. 

Due to existing hardware platforms relying mainly on the Semi-Global Matching (SGM) \cite{hirschmuller2005accurate} or the basic Block Matching stereo algorithms, we will consider these two options in order to distill proxy disparities within \algoname{}++. 

\rev{
\textbf{Proxy Filtering.}
As traditional stereo matchers deliver noisy disparity maps, an effective criterion, $\eta(p)$ in \autoref{eq:eta}, is necessary to filter out outliers and provide reliable supervision to the  continual adaptation process.
In \cite{Tonioni_2017_ICCV,tonioni2019unsupervised} confidence estimation came from a large neural network, which, in our framework, would  add  a substantial computational overhead and prevent continual adaptation in real-time. Thus, in \algoname{}++  we pursue a different approach and rely on computationally efficient strategies geared toward the adopted stereo algorithm. Following \cite{aleotti2020reversing}, we deploy a simple left-right check for SGM and a combination of six confidence measures proposed in \cite{Tosi_2017_BMVC} for Block Matching. From now on, the two settings will be dubbed as SGM and WILD.
}

\begin{figure}[t]
    \centering
    \renewcommand{\tabcolsep}{1pt}
    \begin{tabular}{cc}
        \begin{overpic}[width=0.22\textwidth]{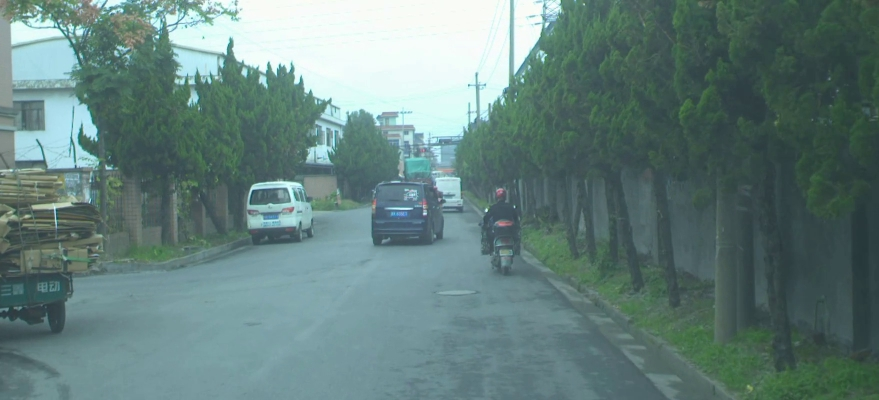} 
        \put (2,38) {$\displaystyle\textcolor{white}{\textbf{(a)}}$}
        \end{overpic} &  
        \begin{overpic}[width=0.22\textwidth]{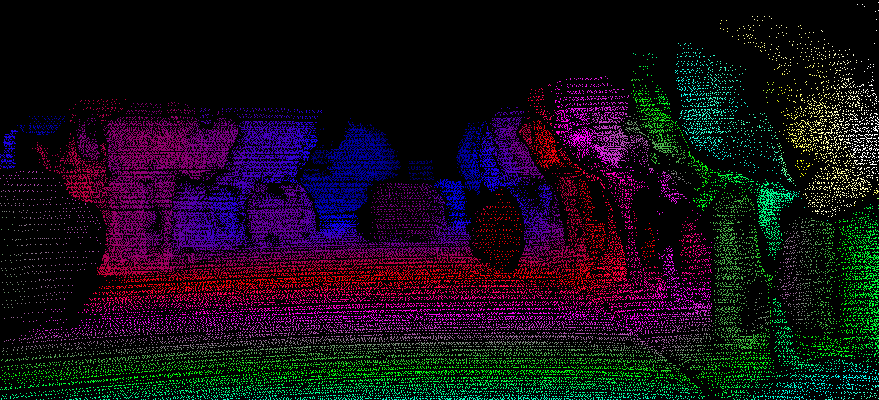}
        \put (2,38) {$\displaystyle\textcolor{white}{\textbf{(b)}}$}
        \end{overpic} \\
        \begin{overpic}[width=0.22\textwidth]{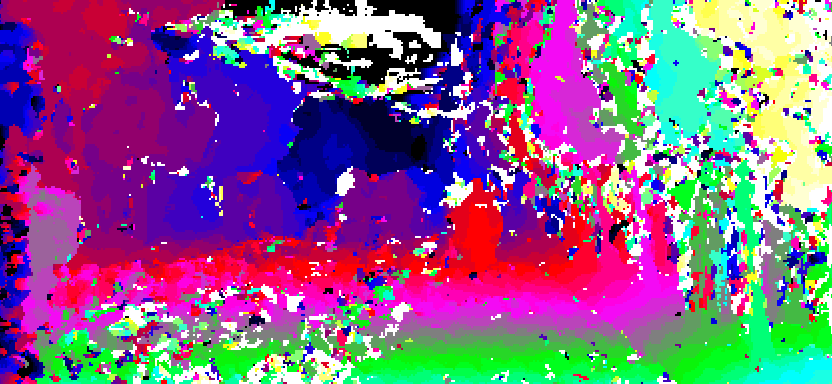} 
        \put (2,38) {$\displaystyle\textcolor{white}{\textbf{(c)}}$}
        \end{overpic} &  
        \begin{overpic}[width=0.22\textwidth]{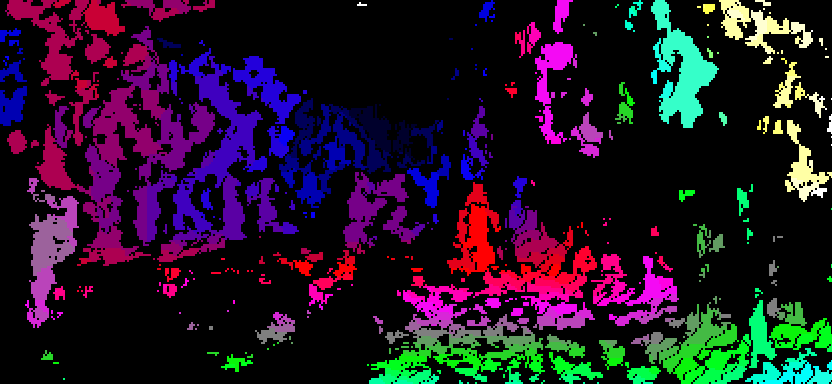}
        \put (2,38) {$\displaystyle\textcolor{white}{\textbf{(d)}}$}
        \end{overpic} \\
        \begin{overpic}[width=0.22\textwidth]{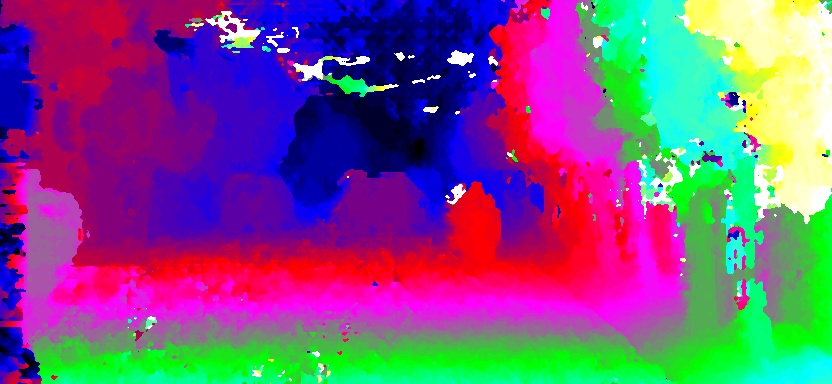} 
        \put (2,38) {$\displaystyle\textcolor{white}{\textbf{(e)}}$}
        \end{overpic} &  
        \begin{overpic}[width=0.22\textwidth]{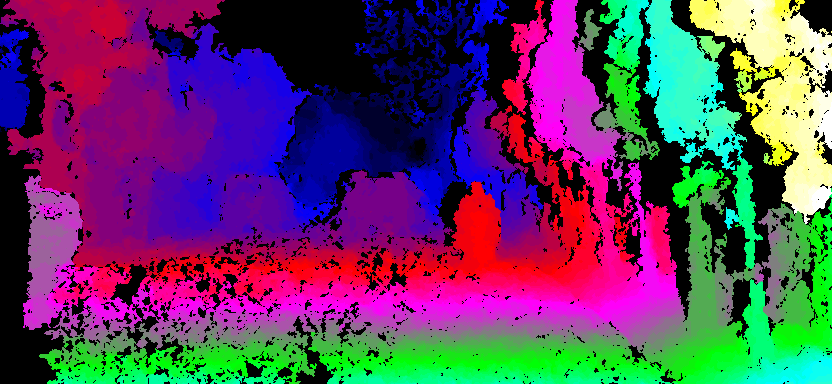}
        \put (2,38) {$\displaystyle\textcolor{white}{\textbf{(f)}}$}
        \end{overpic} \\        
    \end{tabular}
    \caption{\textbf{Proxy labels by the considered stereo pipelines.} The first row depicts a reference image from DrivingStereo  (a)  alongside the available ground-truth disparities (b). The next two rows report the raw disparities and proxy labels (\ie filtered disparities) obtained by the WILD (c),(d) and SGM (e),(f) pipelines. }
    \label{fig:proxies}
\end{figure}

In \autoref{fig:proxies} we show qualitative examples of proxies obtained with the aforementioned pipelines on the DrivingStereo dataset \cite{Yang_2019_CVPR}. We point out that both succeed in providing  reliable proxy labels only, as required by the framework set forth in  \cite{Tonioni_2017_ICCV,tonioni2019unsupervised},  with density depending on the accuracy of the actual stereo matcher.

\begin{table}[t]
\scalebox{0.75}{
\centering
\begin{tabular}{ p{1.2cm}|p{0.8cm} |p{0.8cm} |p{0.8cm} |p{1.5cm}| p{1.6cm} | p{2.0cm}}
    \hline
    \textbf{Layer}  & \textbf{Kernel} & \textbf{Stride} & \textbf{Dilate} & \textbf{Output Channels} & \textbf{Input/Output Resolution} & \textbf{Input}  \\ 
    \hline
    \multicolumn{6}{c}{\textbf{Encoder}} \\
    \hline
    f\_conv1\_a  & 3 & 2 & 1 &  16 & $1/2$ & input\\
    f\_conv1\_b  & 3 & 1 & 1 &  16 & $1/2$  & f\_conv1\_a\\
    f\_conv2\_a  & 3 & 2 & 1 &  32 & $1/{4}$  & f\_conv1\_b\\
    f\_conv2\_b  & 3 & 1 & 1 &  32 & $1/{4}$  & f\_conv2\_a\\
    f\_conv3\_a  & 3 & 2 & 1 &  64 & $1/{8}$  & f\_conv2\_b\\
    f\_conv3\_b  & 3 & 1 & 1 &  64 & $1/{8}$  & f\_conv3\_a\\
    f\_conv4\_a  & 3 & 2 & 1 &  96 & $1/{16}$  & f\_conv3\_b\\
    f\_conv4\_b  & 3 & 1 & 1 &  96 & $1/{16}$  & f\_conv4\_a\\
    f\_conv5\_a  & 3 & 2 & 1 &  128 & $1/{32}$ & f\_conv4\_b\\
    f\_conv5\_b  & 3 & 1 & 1 &  128 & $1/{32}$  & f\_conv5\_a\\
    f\_conv6\_a  & 3 & 2 & 1 & 192  & $1/{64}$  & f\_conv5\_b\\
    f\_conv6\_b  & 3 & 1 & 1 & 192  & $1/{64}$  & f\_conv6\_a\\    
    \hline
    \hline
    \multicolumn{6}{c}{\textbf{Decoder}} \\
    \hline
    d\_conv1  & 3 & 1 & 1 &  128 & 1 & input\\
    d\_conv2  & 3 & 1 & 1 &  128 & 1  & d\_conv1\\
    d\_conv3  & 3 & 1 & 1 &  96 & 1  & d\_conv2\\
    d\_conv4  & 3 & 1 & 1 &  64 & 1  & d\_conv3\\
    d\_conv5  & 3 & 1 & 1 &  32 & 1 & d\_conv4\\
    y  & 3 & 1 & 1 &  1 & 1  & d\_conv5\\
    \hline
    \multicolumn{6}{c}{\textbf{Refinement}} \\
    \hline
    r\_conv1  & 3 & 1 & 1 &  128 & 1 & input\\
    r\_conv2  & 3 & 1 & 2 &  128 & 1  & r\_conv1\\
    r\_conv3  & 3 & 1 & 4 &  128 & 1  & r\_conv2\\
    r\_conv4  & 3 & 1 & 8 &  64 & 1  & r\_conv3\\
    r\_conv5  & 3 & 1 & 16 &  32 & 1 & r\_conv4\\
    r\_conv6  & 3 & 1 & 1 &  1 & 1  & r\_conv5\\
    \hline

\end{tabular}
}
\rev{\caption[Network architecture]{\textbf{ \netname{} architecture.} We detail the layers of the pyramidal features extractor (Encoder),  disparity estimators (Decoder) and refinement module (Refinement).}}
\label{tab:madnet}
\end{table}

\subsection{\extendednetname{} -- \netname{}}
\label{sec:madnet}

In this section we detail the network architecture designed following the abstraction depicted in \autoref{fig:architecture} and deployed throughout the experimental evaluation presented in \autoref{sec:expres}.  
Besides the requirements set forth by modular adaptation (Algo \autoref{algo:mad}), we have developed \extendednetname{} (\netname{}) to achieve a good balance between speed and accuracy. Indeed, \netname{} has a smaller memory footprint and delivers disparity maps much more rapidly with a small loss in accuracy compared to complex networks such as \cite{Kendall_2017_ICCV,chang2018pyramid,liang2018learning}. To design (\netname{}) we took inspiration from recent architectures for optical flow \cite{Ranjan_2017_CVPR,sun2018pwc} and conceived a novel light-weight model for stereo depth estimation. 

We pursue dense disparity regression by a pyramidal approach amenable to modularize the architecture and conducive to fast processing. 
Following \autoref{fig:architecture}, our network estimates disparities $y^i$, with $i \in [1,\dots,5]$, ranging from $\frac{1}{4}$ to $\frac{1}{64}$ resolution, respectively. 
\rev{
As for the encoding section, we design two pyramidal features extractors with shared weights that process the left and right image through a cascade of blocks, as detailed in \autoref{tab:madnet} (\textbf{Encoder}). 
Based on the lowest resolution feature maps, a correlation layer \cite{mayer2016large} computes raw matching costs between the left and right images. Then, we deploy a disparity decoder, detailed in \autoref{tab:madnet} (\textbf{Decoder}), predicting the disparity prediction at the lowest resolution, \ie{} $y^5$. 
}

The lowest resolution disparity map is up-sampled by bilinear interpolation to  2$\times$ the resolution and used to warp the right features towards the left ones, with both feature maps then forwarded to a further correlation layer and another disparity Decoder in order to estimate $y^4$. Again, this map  is up-sampled to the next resolution level and the same computation as in the previous level is carried in order to come up with disparity estimate $y^3$. This scheme is repeated until estimate $y^1$ is reached. \rev{The warping mechanism is instrumental to use a small search range at any resolution level, \ie{} $[-2,2]$, that is equivalent, at the lowest scale, to a search range of 128, to which a further 2$^i$ range is added at each up-sampling. 
The highest resolution disparity estimate, $y^1$, is further processed by a refinement module \cite{sun2018pwc}, as detailed in \autoref{tab:madnet} (\textbf{Refinement}).} 
Finally, the refined $y^1$ is up-sampled from $\frac{1}{4}$ to full resolution by bilinear interpolation.

\begin{table*}[h]
    \centering
    \scalebox{0.85}{
    \begin{tabular}{|c|c|c|c|c|c|c|c|c|cc|c|c|c|c|c|c|c|c|c|c|}
        \cline{1-9}\cline{12-15}
        & LEAStereo & CSPN & GANet & AcfNet & HD$^3$ & DeepPruner & GWCNet & Bi3D & & & AANet & DispNetC & \netname{} & StereoNet  \\
        & \cite{cheng2020hierarchical} & \cite{cheng2019learning} & \cite{Zhang2019GANet} & \cite{zhang2020adaptive} & \cite{yin2019hierarchical} & \cite{Duggal2019ICCV} &  \cite{guo2019group} & \cite{badki2020bi3d} & & &  \cite{xu2020aanet} &  \cite{mayer2016large} & \cite{Tonioni_2019_CVPR} &  \cite{khamis2018stereonet}  \\
        \cline{1-9}\cline{12-15}
        D1-all & 1.65 & 1.74 & 1.81 & 1.89 & 2.02 & 2.15 & 2.11 & 2.21 & \multicolumn{2}{c|}{...} & 2.55 & 4.34 & 4.66 & 4.83 \\
        Time\textdagger{} & 0.30 & 1.00 & 1.80 & 0.18 & 0.48 & 0.14 & 0.32 & 0.48 & & & 0.06 & 0.06 & 0.02 & 0.02 \\
         \cline{1-9}\cline{12-15}
    \end{tabular}
    }
    \vspace{3pt}
    \caption{\rev{\textbf{Comparison between stereo architectures on the \kitti{} 2015 test set without adaptation.} Detailed results available in the KITTI online leader-board. \textdagger{} Times (measured on different GPUs) taken from the online benchmark.}}
    \label{tab:submission}
\end{table*}

With reference to \autoref{fig:architecture}, in \netname{} a generic module $\Theta^i$ consists of a block from the encoding section together with the corresponding disparity decoder, with the arcs linking together the encoder and decoder within a module realized by the warping and correlation layers. Yet, due to $y^1$  being at quarter resolution, $\Theta^1$ is slightly different: in includes  the first two of the six encoding blocks alongside both the disparity decoder and the refinement network. 

\section{Experimental results}
\label{sec:expres}
In this section, we wish to evaluate  thoroughly the effectiveness of our continual adaptation framework. Purposely, we run a set of experiments on a variegated family of datasets.

\begin{table*}[t]
	\center
	\setlength{\tabcolsep}{7pt}
	\scalebox{0.9}
	{
	\begin{tabular}{|l|ll|ll|ll|ll|ll|l|}
		\cline{4-11}
		\multicolumn{3}{c}{} & \multicolumn{2}{|c|}{City (8027 frames)} & \multicolumn{2}{c|}{Residential (28067 frames)} & \multicolumn{2}{c|}{Campus (1149$\times2$ frames)} & \multicolumn{2}{c|}{Road (5674 frames)} \\
		\hline
		Starting Model & Adapt. Mode & Proxy src. & D1-all(\%) & EPE & D1-all(\%) & EPE & D1-all(\%) & EPE & D1-all(\%) & EPE \\
		\hline
		\hline
        \netname{} & No & \xmark & 37.42 & 9.96 & 37.04 & 11.34 & 51.98 & 11.94 & 47.45 & 15.71 \\
        \hline
		\hline
		
		\netname{} & FULL & \xmark & 3.35 & 1.11 & 2.38 & 0.94 & 10.62 & 1.78 & 2.72 & 1.04 \\
        \netname{} & MAD & \xmark & 7.51 & 1.63 & 4.37 & 1.32 & 22.27 & 3.66 & 9.38 & 2.04 \\
        & & & \textcolor{blue}{(+4.16)} & \textcolor{blue}{(+0.52)} & \textcolor{blue}{(+1.99)} & \textcolor{blue}{(+0.37)} & \textcolor{blue}{(+11.65)} & \textcolor{blue}{(+1.88)} & \textcolor{blue}{(+6.66)} & \textcolor{blue}{(+1.00)} \\
		\hline
		
		\netname{} & FULL++ & SGM \cite{hirschmuller2005accurate} & 3.51 & 1.12 & 2.27 & 0.94 & 9.69 & 1.63 & 3.18 & 1.05 \\
        \netname{} & MAD++ & SGM \cite{hirschmuller2005accurate} & 4.12 & 1.18 & 3.31 & 1.04 & 11.24 & 1.76 & 5.32 & 1.22 \\
        & & & \textcolor{blue}{(+0.62)} & \textcolor{blue}{(+0.06)} & \textcolor{blue}{(+1.04)} & \textcolor{blue}{(+0.10)} & \textcolor{blue}{(+1.55)} & \textcolor{blue}{(+0.13)} & \textcolor{blue}{(+2.14)} & \textcolor{blue}{(+0.17)} \\
 
		\hline
	    \netname{} & FULL++ & WILD \cite{Tosi_2017_BMVC} & 5.11 & 1.23 & 2.82 & 0.99 & 11.79 & 1.89 & 4.28 & 1.11 \\
        \netname{} & MAD++ & WILD \cite{Tosi_2017_BMVC} & 5.75 & 1.30 & 2.88 & 0.99 & 13.93 & 2.04 & 5.39 & 1.24 \\
        & & & \textcolor{blue}{(+0.65)} & \textcolor{blue}{(+0.07)} & \textcolor{blue}{(+0.06)} & \textcolor{blue}{(-0.01)} & \textcolor{blue}{(+2.13)} & \textcolor{blue}{(+0.15)} & \textcolor{blue}{(+1.11)} & \textcolor{blue}{(+0.13)} \\
		\hline
	\end{tabular}
	}
	\vspace{1pt}
	\caption{\textbf{Online adaptation within a single domain}. Results on  the \emph{City}, \emph{Residential}, \emph{Campus} and \emph{Road} sequences from \kitti{} \cite{KITTI_RAW}.}		
	\label{tab:kitti_sequences}
\end{table*}

\subsection{Datasets}
\label{sec:datasets}

Here, we provide a description of the datasets used for the experiments.

\textbf{FlyingThings3D.} A collection of synthetic images, made out of about 22k training stereo pairs with dense ground-truth labels, part of the SceneFlow synthetic dataset \cite{mayer2016large}. This dataset has been used to pre-train \netname{} before deployment on real images, according to the standard practice in recent deep stereo literature outlined in \cite{Tonioni_2019_CVPR}. \rev{Specifically, we perform 1.2M training iterations using Adam Optimizer and a learning rate of 10$^{-4}$, halved after 400K steps and further every 200K until convergence. 
As loss function, we compute the L1 difference between disparity maps estimated at each resolution and downsampled ground-truth labels.
The final loss is a weighted sum of the contributions from the different resolutions, with weights set to  0.005, 0.01, 0.02, 0.08, 0.32 from $y^2$ to $y^6$ according to  \cite{sun2018pwc}.}

\textbf{KITTI 2015 train set.} A collection 200 stereo pairs with sparse ground-truth maps, obtained from post-processed LIDAR measurements and 3D CAD objects \cite{KITTI_2015}. This dataset has been used to fine-tune \netname{} \rev{for 500K steps with learning rate $10^{-4}$ by computing the loss only on the full-resolution disparity map and using 0.001 as weight, as described in \cite{Tonioni_2019_CVPR}. From now on, a \netname{} model trained on FlyingThings3D and fine-tuned on KITTI 2015 with explicit supervision will be referred to as \netname{}-GT.}

\textbf{Raw \kitti{}.} A large dataset featuring 61 stereo sequences, for a total of about 43k pairs with different image resolution. We use  a constant resolution of $320\times1226$ pixels by taking  central crops of the original frames \cite{Tonioni_2019_CVPR}. As depth ground-truths, we use filtered LIDAR measurements \cite{Uhrig2017THREEDV} converted to disparities through  known calibration parameters \cite{Tonioni_2019_CVPR}. According to the classification reported in the official website, we group  sequences into four main categories: \textit{Road}, \textit{Residential}, \textit{Campus} and \textit{City}. Then, we concatenate the sequences belonging to the same category so as to obtain new, longer sequences of 5674, 28067, 1149$\times2$\footnote{About \textit{Campus}, it represents the most challenging environment characterized by low-textured buildings, yet the shortest sequence. For this reason, we loop twice over the sequence as in \cite{Tonioni_2019_CVPR}.} and 8027 frames for the above mentioned categories, respectively.
In this manner, we simulate four macro environments characterised by different peculiarities, \ie{} \textit{City} and \textit{Residential} mostly show roads surrounded by buildings, while \textit{Road} images are collected while driving in highways and country roads, thus mainly depicting cars and vegetation. The dataset  provides also raw LIDAR measurements.

\textbf{DrivingStereo.} A recent dataset \cite{Yang_2019_CVPR} collecting about 170k stereo images grouped in 38 sequences with average resolution of $384 \times 832$ pixels. ground-truth is obtained by iterative filtering of LIDAR labels by means of a stereo CNN. We select three challenging sequences, namely  \textit{2018-08-17-09-45}, \textit{2018-10-11-17-08} and \textit{2018-10-15-11-43}, consisting of 1667, 1119 and 4950 frames, respectively. We rename the above mentioned sequences as   \textit{Rainy}, \textit{Cloudy} and \textit{Country}, respectively, according to their main peculiarities, We selected these sequences to 1) evaluate short-term adaptation (\ie{}, after few hundreds frames) in challenging conditions (\eg{}, rainy) and 2) assess the impact of prior continual adaptation (\eg{}, on \kitti) when moving to a new environment.

\textbf{WeanHall.} An indoor dataset \cite{weanhall} which includes 6510 stereo pairs. As the working environment is very different from the autonomous driving scenarios addressed by previous datasets, we deem it worth evaluating performance also when continually adapting \netname{} in so diverse settings.
Unfortunately, no ground-truth disparities/depths are provided in WeanHall, neither are we aware of any other indoor stereo dataset providing sequences of real images alongside with the corresponding ground-truth labels. Therefore, we rely on the  photometric error (\autoref{eq:photo}) to provide quantitative performace figures on WeanHall.

\subsection{Experimental protocol}

To assess the performance of our continual adaptation schemes, we run disparity prediction on the stereo pairs of a given sequence according to their order, \ie{} as if they were acquired online in the field. On \kitti{} and DrivingStereo we measure the D1-all error rate as the percentage of pixels having absolute disparity error larger than 3 and relative error larger than 5\%, as well as the End-Point-Error (EPE), whilst on WeanHall we measure the photometric error upon reprojection, as detailed in \autoref{eq:photo}.
In case of baseline performance dealing with prediction without adaptation, we simply compute error metrics for each stereo frame  and average them across the entire sequence. In case of adaptation, we process stereo pair $x_t$ to predict a disparity map $y_t$ and compute the error metrics on it, then we update the network according to either Algo \autoref{algo:backprop} (FULL) or Algo \autoref{algo:mad} (\ie MAD or \algoname{}++). As a consequence, the impact of the continual adaptation step at time $t$ will affect the error metrics from time $(t+1)$. Akin to baseline performance, per-sequence metrics are computed by averaging across frames those dealing with the per-frame predictions $y_t$. 
\rev{During adaptation, we use a momentum optimizer, with a constant learning rate of 10$^{-4}$ and a momentum of 0.9. Different optimizers or learning rate schedules resulted in minor fluctuations in our experiments.}
All experiments have been carried out using the source code and trained models available at \url{github.com/CVLAB-Unibo/Real-time-self-adaptive-deep-stereo}.

\rev{
\subsection{Comparison with state-of-the-art stereo networks}

We start by comparing  \netname{} to the current state-of-the-art networks on the standard KITTI 2015 benchmark setting, \ie{} without performing online adaptation. \autoref{tab:submission} collects performance figures retrieved from the online benchmark (to which we refer the reader for a complete overview) for both accuracy and runtime, although the latter is measured on GPUs belonging to different hardware generations. 
We can notice that the most accurate architectures trade precision for speed, whereas \netname{} can operate in real-time with a modest increase of the error rate (about 3\% with respect to the top-performing network). The focus of our work concern deploying\netname{} alongside suitable online adaptation strategies to maximize accuracy while running in real-time in unknown environments.
}

\subsection{Evaluation on \kitti{}} \label{sec:kitti}

We begin our evaluation by studying different aspects of continual adaptation on the KITTI dataset. First, we address short-term adaptation within a domain by considering the KITTI sequences belonging to the same category independently. Then, we tackle a setup dealing with long-term adaptation across domains by concatenating together the sequences belonging to the different categories. We report results obtained by both the continual adaptation schemes discussed in \autoref{sec:continual}, namely 
full adaptation (Algo \autoref{algo:backprop}) and the more efficient - though approximated- modular adaptation approach (Algo \autoref{algo:mad}). We also assess upon steering both adaptation schemes by either self-supervision of proxy supervision. As for the latter source of supervision, we consider proxy labels yielded by the previously described SGM and WILD pipelines. As in KITTI raw LIDAR measurements are available alongside stereo pairs, we also investigate on the effectiveness of this form of proxy supervision.
Finally, we dig deeper into our framework by analysing the distribution of the update steps across the modules in modular adaptation (Algo \autoref{algo:mad} and investigating on the computational savings that may be achieved by employing slower adaptation rates. 

\begin{table}[t]
	\setlength{\tabcolsep}{9pt}
	\center
    \scalebox{0.82}
    {
			\begin{tabular}{|l|ll|ll|l|}
				\hline
				Starting Model & Adapt. Mode & Proxy src. & D1-all(\%) & EPE \\
				\hline
				\netname{} & No & \xmark & 38.84 & 11.68 \\
				\hline
				\netname{} & FULL & \xmark & 2.43 & 0.95 \\
                \netname{} & MAD & \xmark & 4.09 & 1.19 \\
                & & & \textcolor{blue}{(+1.66)} & \textcolor{blue}{(+0.24)} \\
				\hline
				
				\netname{} & FULL++ & SGM \cite{hirschmuller2005accurate} & 2.28 & 0.95 \\
                \netname{} & MAD++ & SGM \cite{hirschmuller2005accurate} & 2.46 & 0.98 \\
                & & & \textcolor{blue}{(+0.18)} & \textcolor{blue}{(+0.03)} \\
				\hline
				
				\netname{} & FULL++ & WILD \cite{Tosi_2017_BMVC} & 2.64 & 0.98 \\
                \netname{} & MAD++ & WILD \cite{Tosi_2017_BMVC} & 2.44 & 0.96 \\
                & & & \textcolor{blue}{(-0.20)} & \textcolor{blue}{(-0.02)} \\
				 \hline
	            \end{tabular}
	}
	\vspace{1pt}
	\caption{\textbf{Online adaptation across different domains.} Results on the sequence \emph{Campus} $\rightarrow$ \emph{City} $\rightarrow$ \emph{Residential} $\rightarrow $ \emph{Road} (\about 43k frames, the whole KITTI dataset)}	
	
	\label{tab:overall}
\end{table}

\begin{figure}[t]
    \centering
    \renewcommand{\tabcolsep}{1pt}
    \begin{tabular}{c}
    \includegraphics[width=0.48\textwidth]{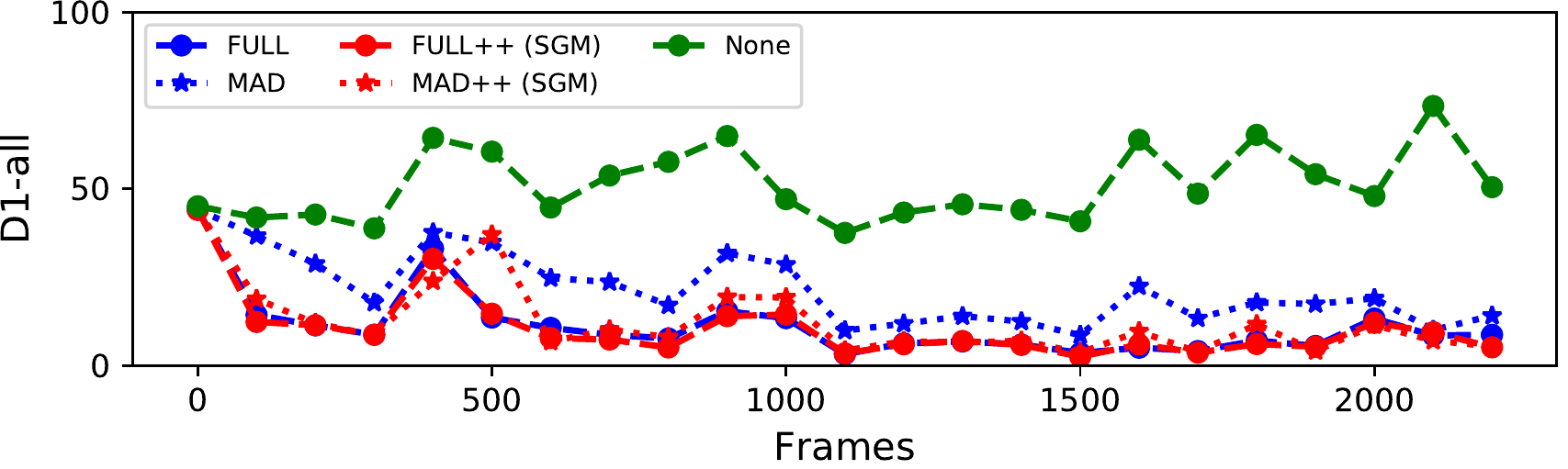} \\    \end{tabular}
    \caption{\textbf{Adaptation speed on \textit{Campus}.} \algoname{}++ adapts much faster than \algoname{}, rapidly converging to the same error level as FULL and FULL++ (blue and red solid lines, almost completely overlapped). 
    }
    \label{fig:curves}
\end{figure}

\begin{figure}[t]
    \centering
    \renewcommand{\tabcolsep}{1pt}
    \begin{tabular}{cc}
    \begin{overpic}[height=0.07\textwidth]{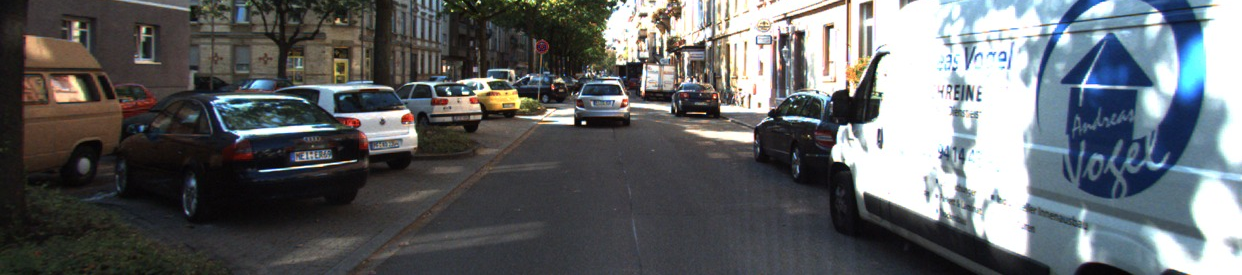} 
    \put (1,2) {$\displaystyle\textcolor{white}{\textbf{(a)}}$}
    \end{overpic} &
    \includegraphics[height=0.07\textwidth]{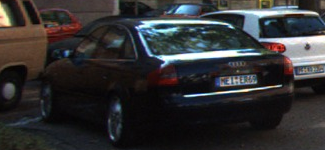}
    \\
    \begin{overpic}[height=0.07\textwidth]{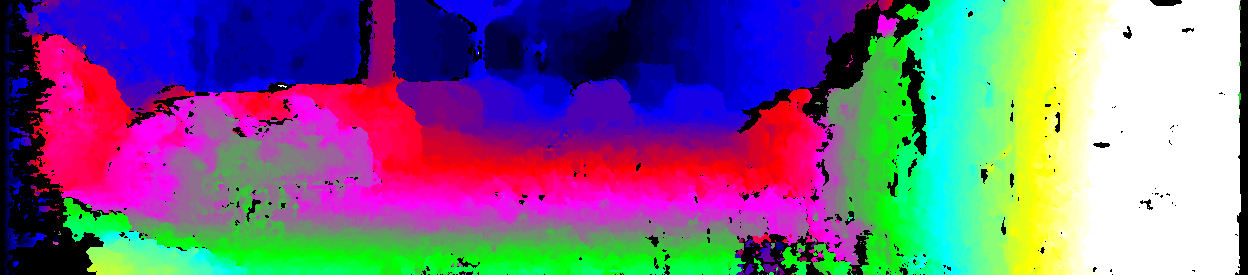} 
    \put (1,2) {$\displaystyle\textcolor{white}{\textbf{(b)}}$}
    \end{overpic} &
    \includegraphics[height=0.07\textwidth]{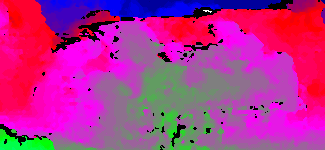}
    \\
    \begin{overpic}[height=0.07\textwidth]{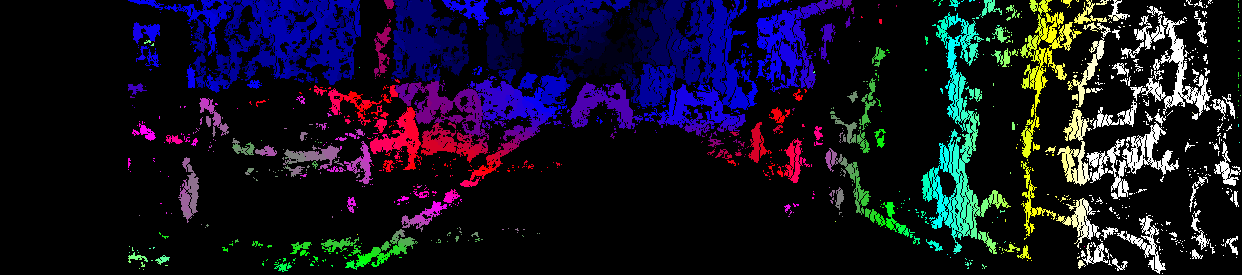} 
    \put (1,2) {$\displaystyle\textcolor{white}{\textbf{(c)}}$}
    \end{overpic} &
    \includegraphics[height=0.07\textwidth]{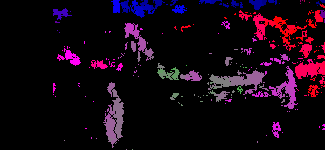}
    \\
    \end{tabular}
    \caption{\textbf{Comparison between different proxy labels.} We show a reference image (a) from the \textit{City} domain and proxy labels sourced by SGM (b) and WILD (c).}
    \label{fig:density}
\end{figure}

\begin{table}[t]
	\center
	\setlength{\tabcolsep}{7pt}
	\scalebox{0.9}
	{
	\begin{tabular}{|l|l||ll|}
		\hline
		Starting Model & Adapt. Mode & D1-all(\%) & EPE \\
		\hline
		\hline
		\netname{}-Synthia & FULL (baseline) & 2.76 & 1.03 \\
		\netname{}-Synthia-FOL2A \cite{Tonioni_2019_learn2adapt} & FULL & 2.60 & 1.01 \\
        \hline
        \netname{}-Synthia & FULL++ (ours) & 2.23 & 0.95 \\
        \hline
	\end{tabular}
	}
	\vspace{3pt}
	\rev{\caption{\textbf{Comparison of online adaptation strategies across different domains}. Results on the sequence \emph{Campus} $\rightarrow$ \emph{City} $\rightarrow$ \emph{Residential} $\rightarrow $ \emph{Road}. Comparison between \netname{} trained on FlyingThings3D and finetuned on Synthia, with (FOL2A) or without meta-learning optimization.}}		
	\label{tab:l2a}
\end{table}

\textbf{Online adaptation \textit{in-the-wild}.}
We address here the reference scenario concerning practical deployment \textit{in-the-wild}: a stereo network pre-trained on synthetic data is run in a wholly unknown environment. In this setting, the experimental results reported in \autoref{tab:kitti_sequences}
and \autoref{tab:overall} deals with adaptation within a single domain and across different domains, respectively,  both tables following the same organization. The first row highlights the baseline performance measured by pre-training \netname{} on FlyingThings3D and then running  the model without any kind of online adaptation. Then, we report the results achieved by the two online adaptation schemes, \ie{} FULL (Algo \autoref{algo:backprop}) and  \algoname{} (Algo \autoref{algo:mad}), realized by self-supervision through the photometric error loss, as formulated in our previous work \cite{Tonioni_2019_CVPR}. The next rows concern  the novel formulation described in this paper, which exploits different sources of proxy supervision, namely SGM and WILD. Again, FULL++ and \algoname{}++ refer to  Algo \autoref{algo:backprop} and  Algo \autoref{algo:mad}, respectively. For each source of supervision, we report in blue the difference in terms of D1-all and EPE between the \algoname{} and FULL adaptation schemes, so as to highlight the gap between modular and full adaptation when relying  on the same form of supervision. The difference between the two Tables deals with the KITTI sequences belonging to the same category having been processed individually in \autoref{tab:kitti_sequences}, after concatenation according to the order \emph{Campus} $\rightarrow$ \emph{City} $\rightarrow$ \emph{Residential}  $\rightarrow $ \emph{Road}, in \autoref{tab:overall}.

\begin{table*}[t]
	\center
	\setlength{\tabcolsep}{7pt}
	\scalebox{0.9}
	{
	\begin{tabular}{|l|ll|ll|ll|ll|ll|l|}
		\cline{4-11}
		\multicolumn{3}{c}{} & \multicolumn{2}{|c|}{City (8027 frames)} & \multicolumn{2}{c|}{Residential (28067 frames)} & \multicolumn{2}{c|}{Campus (1149$\times2$ frames)} & \multicolumn{2}{c|}{Road (5674 frames)} \\
		\hline
		Starting Model & Adapt. Mode & Proxy src. & D1-all(\%) & EPE & D1-all(\%) & EPE & D1-all(\%) & EPE & D1-all(\%) & EPE \\
		\hline
		\hline
		\netname{}-GT & No & \xmark & 2.08 & 0.80 & 2.55 & 0.91 & 6.51 & 1.31 & 1.63 & 0.83 \\
		\hline
		\hline
		
		\netname{}-GT & FULL & \xmark & 1.60 & 0.89 & 1.87 & 0.86 & 4.70 & 1.31 & 1.14 & 0.81 \\
        \netname{}-GT & MAD & \xmark & 1.62 & 0.90 & 1.54 & 0.85 & 4.81 & 1.28 & 1.28 & 0.84 \\
        & & & \textcolor{blue}{(+0.02)} & \textcolor{blue}{(+0.01)} & \textcolor{blue}{(-0.33)} & \textcolor{blue}{(-0.01)} & \textcolor{blue}{(+0.10)} & \textcolor{blue}{(-0.03)} & \textcolor{blue}{(+0.14)} & \textcolor{blue}{(+0.03)} \\
		\hline
		
		\netname{}-GT & FULL++ & SGM \cite{hirschmuller2005accurate} & 1.59 & 0.92 & 1.49 & 0.84 & 3.50 & 1.05 & 1.24 & 0.85 \\
        \netname{}-GT & MAD++ & SGM \cite{hirschmuller2005accurate} & 1.59 & 0.92 & 1.52 & 0.86 & 3.73 & 1.14 & 1.35 & 0.86 \\
        & & & \textcolor{blue}{(-0.01)} & \textcolor{blue}{(0.00)} & \textcolor{blue}{(+0.03)} & \textcolor{blue}{(+0.02)} & \textcolor{blue}{(+0.22)} & \textcolor{blue}{(+0.09)} & \textcolor{blue}{(+0.11)} & \textcolor{blue}{(+0.01)} \\
		 \hline
		 
		\netname{}-GT & FULL++ & WILD \cite{Tosi_2017_BMVC} & 1.58 & 0.90 & 1.50 & 0.85 & 4.19 & 1.13 & 1.24 & 0.83 \\
        \netname{}-GT & MAD++ & WILD \cite{Tosi_2017_BMVC} & 1.57 & 0.91 & 1.79 & 0.87 & 4.26 & 1.23 & 1.30 & 0.85 \\
        & & & \textcolor{blue}{(-0.01)} & \textcolor{blue}{(+0.01)} & \textcolor{blue}{(+0.29)} & \textcolor{blue}{(+0.02)} & \textcolor{blue}{(+0.07)} & \textcolor{blue}{(+0.10)} & \textcolor{blue}{(+0.06)} & \textcolor{blue}{(+0.02)} \\
		\hline
	\end{tabular}
	}
	\vspace{1pt}
	\caption{\textbf{Online adaptation within a single domain after fine-tuning}. Results on  the \emph{City}, \emph{Residential}, \emph{Campus} and \emph{Road} sequences from \kitti{} \cite{KITTI_RAW}. \emph{-GT} denotes fine-tuning by ground-truths on the \kitti{} training set after pre-training on synthetic imagery.}		
	\label{tab:kitti_sequences_gt}
\end{table*}

\begin{table}[t]
	\setlength{\tabcolsep}{9pt}
	\center
    \scalebox{0.82}
    {
			\begin{tabular}{|l|ll|ll|l|}
				\hline
				Starting Model & Adapt. Mode & Proxy src. & D1-all(\%) & EPE \\
				\hline
				\netname{}-GT & No & \xmark & 2.45 & 0.89 \\
				\hline
				\netname{}-GT & FULL & \xmark & 1.83 & 0.88 \\
                \netname{}-GT & MAD & \xmark & 1.94 & 0.86 \\
                & & & \textcolor{blue}{(+0.11)} & \textcolor{blue}{(-0.01)} \\
				\hline
				
				\netname{}-GT & FULL++ & SGM \cite{hirschmuller2005accurate} & 1.46 & 0.85 \\
                \netname{}-GT & MAD++ & SGM \cite{hirschmuller2005accurate} & 1.76 & 0.89 \\
                & & & \textcolor{blue}{(+0.30)} & \textcolor{blue}{(+0.03)} \\
				\hline
				
				\netname{}-GT & FULL++ & WILD \cite{Tosi_2017_BMVC} & 1.48 & 0.85 \\
                \netname{}-GT & MAD++ & WILD \cite{Tosi_2017_BMVC} & 1.64 & 0.86 \\
                & & & \textcolor{blue}{(+0.16)} & \textcolor{blue}{(+0.01)} \\
				\hline
	\end{tabular}
	}
	\vspace{1pt}
	\caption{\textbf{Online adaptation across different domains after fine-tuning.} Results on the sequence \emph{Campus} $\rightarrow$ \emph{City} $\rightarrow$ \emph{Residential} $\rightarrow $ \emph{Road}. \emph{-GT} denotes fine-tuning by ground-truths on the \kitti{} training set after pre-training on synthetic imagery.}	
	\label{tab:overall_gt}
\end{table}

Firstly, the comparison between the baseline performance reported in the first row of both Tables and the figures in the successive ones highlights the dramatic error drops yielded by all considered methods and vouches for 
the utmost effectiveness of adapting online a stereo model pre-trained on synthetic data and run in a wholly unknown environment. 
As expected, between the two schemes, full adaptation  (FULL/FULL++) consistently outperforms modular adaptation (\algoname{}/\algoname{}++) when steered by the same kind of supervision. Then, as for the former scheme, self-supervision (FULL) and proxy supervision (FULL++) seem, overall,  rather equivalent options, one or the other performing slightly better in some experiments: e.g., in \autoref{tab:kitti_sequences} FULL provides the lowest  D1-all error in City and Road, FULL++(SGM) in Residential and Campus, whereas in \autoref{tab:overall} FULL++(SGM) yields smaller errors than FULL which, in turn, outperforms FULL++(WILD). 

However, when dealing with modular adaptation, proxy supervision (\algoname{}++) consistently outperforms self-supervision (\algoname{}), often by a very large margin. Indeed, unlike self-supervision, proxy supervision allows for reducing the performance gap between full and modular adaptation dramatically, as highlighted by the figures reported in blue in \autoref{tab:kitti_sequences}.
We also point out that this is particularly evident in short sequences, such as \textit{Campus} in \autoref{tab:kitti_sequences}, for which the gap is reduced from about 11.83\% to less than 3\%. Indeed, as shown in \autoref{fig:curves}, \algoname{}++ is much faster than \algoname{} in reaching the same accuracy level as FULL/FULL++, which suggests performance differences measured by error metrics averaged along a sequence likely showing up more evidently in shorter ones. Similar to \autoref{tab:kitti_sequences}, proxy supervision turns out particularly beneficial to improve modular adaptation with respect to the formulation based on self-supervision  in the long-term, cross-domain adaptation experiments considered in \autoref{tab:overall}, with \algoname{}++ turning out almost as effective as FULL++ while observing a substantial gap between MAD and FULL. 

When it comes to reasoning on the  different proxy sources adopted with either FULL++ or \algoname{}++, we can notice that, more often than not, SGM delivers a more effective supervision than WILD. In fact, the latter provides better performance only with MAD++ in the Residential domain (\autoref{tab:kitti_sequences}) and, though rather slightly, in case of cross-domain adaptation ((\autoref{tab:overall}). We ascribe this to the much higher density of proxy labels featured by the SGM pipeline compared to WILD, as illustrated qualitatively in \autoref{fig:density}.

\begin{table*}[t]
	\center
	\setlength{\tabcolsep}{7pt}
	\scalebox{0.9}
	{
	\begin{tabular}{|l|ll|ll|ll|ll|ll|l|}
		\cline{4-11}
		\multicolumn{3}{c}{} & \multicolumn{2}{|c|}{City (8027 frames)} & \multicolumn{2}{c|}{Residential (28067 frames)} & \multicolumn{2}{c|}{Campus (1149$\times2$ frames)} & \multicolumn{2}{c|}{Road (5674 frames)} \\
		\hline
		Starting Model & Adapt. Mode & Proxy src. & D1-all(\%) & EPE & D1-all(\%) & EPE & D1-all(\%) & EPE & D1-all(\%) & EPE \\
		\hline
		\hline
        
        \netname{} & No & \xmark & 37.42 & 9.96 & 37.04 & 11.34 & 51.98 & 11.94 & 47.45 & 15.71 \\
        \hline
		\netname{} & FULL++ & LIDAR & 3.66 & 0.97 & 2.94 & 0.89 & 9.10 & 1.54 & 3.24 & 0.93 \\
        \netname{} & MAD++ & LIDAR & 4.63 & 1.12 & 3.99 & 1.06 & 19.33 & 2.32 & 4.74 & 1.12 \\
        & & & \textcolor{blue}{(+0.97)} & \textcolor{blue}{(+0.15)} & \textcolor{blue}{(+1.05)} & \textcolor{blue}{(+0.18)} & \textcolor{blue}{(+10.23)} & \textcolor{blue}{(+0.78)} & \textcolor{blue}{(+1.50)} & \textcolor{blue}{(+0.20)} \\
		\hline
        \hline

		\netname{}-GT & No & \xmark & 2.08 & 0.80 & 2.55 & 0.91 & 6.51 & 1.31 & 1.63 & 0.83 \\
		\hline
		\netname{}-GT & FULL++ & LIDAR & 2.00 & 0.67 & 2.17 & 0.75 & 4.29 & 0.96 & 1.59 & 0.66 \\
        \netname{}-GT & MAD++ & LIDAR & 3.16 & 0.88 & 2.86 & 0.92 & 4.96 & 1.20 & 1.89 & 0.79 \\
        & & & \textcolor{blue}{(+1.15)} & \textcolor{blue}{(+0.21)} & \textcolor{blue}{(+0.69)} & \textcolor{blue}{(+0.17)} & \textcolor{blue}{(+0.67)} & \textcolor{blue}{(+0.23)} & \textcolor{blue}{(+0.30)} & \textcolor{blue}{(+0.14)} \\
		\hline
	\end{tabular}
	}
	\vspace{1pt}
	\caption{\textbf{Online adaptation within a single domain with proxy supervision from raw LIDAR.} Results on  the \emph{City}, \emph{Residential}, \emph{Campus} and \emph{Road} sequences from \kitti{} \cite{KITTI_RAW}. \emph{-GT} denotes fine-tuning by ground-truths on the \kitti{} training set after pre-training on synthetic imagery.}		
	\label{tab:kitti_sequences_lidar}
\end{table*}

\begin{table}[t]
	\setlength{\tabcolsep}{9pt}
	\center
    \scalebox{0.82}
    {
			\begin{tabular}{|l|ll|ll|l|}
				\hline
				Starting Model & Adapt. Mode & Proxy src. & D1-all(\%) & EPE \\
				\hline
				\netname{} & No & \xmark & 38.84 & 11.68 \\
				\hline
				\netname{} & FULL++ & LIDAR & 2.87 & 0.87 \\
                \netname{} & MAD++ & LIDAR & 3.86 & 1.02 \\
                & & & \textcolor{blue}{(+0.99)} & \textcolor{blue}{(+0.15)} \\
                \hline
				\hline
				\netname{}-GT & No & \xmark & 2.45 & 0.89 \\
				\hline
				\netname{}-GT & FULL++ & LIDAR & 2.06 & 0.73 \\
                \netname{}-GT & MAD++ & LIDAR & 2.86 & 0.89 \\
                & & & \textcolor{blue}{(+0.81)} & \textcolor{blue}{(+0.16)} \\
                \hline
	            \end{tabular}
	}
	\vspace{1pt}
	\caption{\textbf{Online adaptation across different domains with proxy supervision from raw LIDAR.} Results on the sequence \emph{Campus} $\rightarrow$ \emph{City} $\rightarrow$ \emph{Residential} $\rightarrow $ \emph{Road}. \emph{-GT} denotes fine-tuning by ground-truths on the \kitti{} training set after pre-training on synthetic imagery.}	
	
	\label{tab:overall_lidar}
\end{table}

\rev{
\textbf{Proxy supervision vs meta-learning.} We compare proxy supervision to the meta-learning framework proposed in \cite{Tonioni_2019_learn2adapt}. Purposely, according to the setting suggested in \cite{Tonioni_2019_learn2adapt}, we take the \netname{} model trained on FlyingThings3D and 1) fine-tune it on Synthia in a traditional, supervised manner or 2) using the First Order approximation  variant of L2A (FOL2A).
Thus, for this experiment, 1) provides the baseline with respect to which we evaluate the adaptation performance provided by FOL2A and our proposed approach dealing with proxy labels gathered by SGM. 
In both cases, we use a learning rate of 10$^{-5}$ and a batch size of 16 for 10K steps. For meta-learning optimization we choose FOL2A \cite{Tonioni_2019_learn2adapt}, since all other strategies resulted unstable when coupled with \netname{}, and use 3 consecutive frames.
\autoref{tab:l2a} collects the outcome of this experiment. We can notice how using L2A during pre-training can improve the performance with respect to the baseline. However, our proposal turns out significantly more effective, \ie  proxy supervision is more effective than L2A pre-training. A thorough comparison dealing with all the individual sequences is reported in the supplementary material.}

\textbf{Online adaptation after fine-tuning.}
As proposed in \cite{Tonioni_2019_CVPR}, we also investigate on the effectiveness of the different on-line adaptation schemes in case the pre-trained model may be fine-tuned by real stereo pairs with ground-truths before running inference. Indeed, this is the case of some research datasets like KITTI. 

Thus, in \autoref{tab:kitti_sequences_gt} and \autoref{tab:overall_gt} we report the results dealing with adaptation on each of the four KITTI domains and across domains, the only difference with respect to \autoref{tab:kitti_sequences}, \autoref{tab:overall} being that now  \netname{} has been pre-trained on FlyingThings3D and then fine-tuned on the KITTI 2015 training set before undergoing online adaptation by the considered schemes and forms of supervision.

The first row in \autoref{tab:kitti_sequences_gt} and \autoref{tab:overall_gt} show the baseline performance yielded by running the pre-trained and fine-tuned model with online adaptation turned off. As now the network has been fine-tuned by samples endowed with ground-truth disparities, the baseline model performs considerably better than in \autoref{tab:kitti_sequences} and \autoref{tab:overall}.  Nevertheless, both  full adaptation (FULL/FULL++) as well as modular adaptation (\algoname{}/\algoname{}++) allow to further improve over this strong baseline in all the considered experiments,  typically yielding substantial relative performance gains. For instance, in \autoref{tab:kitti_sequences_gt}, FULL++(SGM) and \algoname{}++(SGM) can provide a relative D1-all error reduction of about 46\% and 42\% on Campus (the shortest sequence), whereas FULL can decrease such error by about 30\% in Road and 
 \algoname{}++(WILD) by about 24\% in City. Similarly, in \autoref{tab:overall_gt}, the relative D1-all error reduction ranges from about 20\% (MAD) to as much as 40\% (FULL++(SGM)). 
 
Besides, due to the base model undergoing adaptation being stronger,  full and modular adaptation tend to exhibit a much smaller gap  when driven by the same form of supervision (figures in blue).  Moreover, unlike the previous experiment, in most cases proxy supervision  provides better performance than self-supervision non only with modular adaptation (\algoname{}++ vs. \algoname{}) but also with full adaptation (FULL++ vs FULL).

As for the two kinds of proxy labels, the WILD pipeline seems now competitive with respect to SGM, as it can provide better or equivalent performance also in case of full adaptation (City and Road in  \autoref{tab:kitti_sequences_gt}) and turns out generally more effective when the model undergoes modular adaptation, in particular in the long-term, cross-domain experiment  (\autoref{tab:overall_gt}). 
We would be led to ascribe this finding to the fact that, although fewer in number, the proxies extracted by WILD are more accurate and thus more amenable to refine the already good disparities predicted by a strong base model fine-tuned by real images equipped with ground-truths. Conversely, as observed in the previous experiment, denser proxies seem instrumental to break down the gross errors spread throughout the image delivered by a baseline prone to the synthetic-to-real domain shift.

\textbf{Proxy supervision by LIDAR}. We also inquire about the effectiveness of continual adaptation in case  proxy supervision may be obtained from raw measurements provided by a LIDAR sensor, as it is the case of the KITTI dataset.  Akin to previous experiments, we consider both a baseline \netname{} pre-trained on FlyingThings3D as well as a model further fine-tuned by labelled stereo pairs from the KITTI 2015 training set (\netname{}-GT). \autoref{tab:kitti_sequences_lidar} and \autoref{tab:overall_lidar} collect the results dealing with online adaptation  on each of the four KITTI domains and across them, respectively.  
We can notice a trend similar to previous experiments relatively to several key findings. Indeed, in the  \textit{in-the-wild} scenario, online adaptation does matter a lot as in both Tables we observe a dramatic reduction of errors compared to the baseline model pre-trained on synthetic imagery. Besides, FULL++ consistently outperform \algoname{}++, the margin turning out generally higher than with proxy supervision by SGM and WILD (\autoref{tab:kitti_sequences} and \autoref{tab:overall}). In case stereo pairs with ground-truth are available to fine-tune the pre-trained model, online adaptation by LIDAR proxies and full adaptation (FULL++) is still beneficial, whilst modular adaptation (\algoname{}++) tend to perform worse than the baseline. Hence, we are lead to conclude that modular adaptation (Algo \autoref{algo:mad}) with  supervision by LIDAR is less effective than in case the proxy labels are gathered by the SGM and WILD pipeline.

\begin{figure}[t]
    \centering
    \renewcommand{\tabcolsep}{1pt}
    \begin{tabular}{cc}
    \begin{overpic}[height=0.07\textwidth]{images/density/left.png} 
    \put (1,2) {$\displaystyle\textcolor{white}{\textbf{(a)}}$}
    \end{overpic} &
    \includegraphics[height=0.07\textwidth]{images/density/left_crop.png}
    \\
    \begin{overpic}[height=0.07\textwidth]{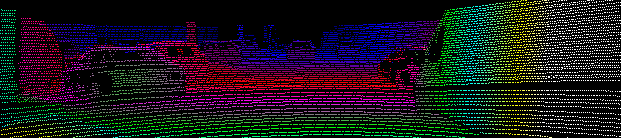} 
    \put (1,2) {$\displaystyle\textcolor{white}{\textbf{(b)}}$}
    \end{overpic} &
    \includegraphics[height=0.07\textwidth]{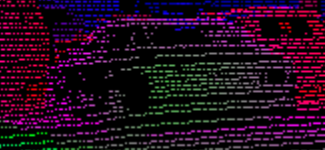}
    \\
    \end{tabular}
    \caption{\textbf{Raw LIDAR for proxy supervision.} We show a reference image (a) from the \textit{City} domain alongside proxy labels sourced by raw Lidar (b), with the latter exhibiting wrong measurements at depth boundaries.}
    \label{fig:lidar_density}
\end{figure}

Moreover, by comparing \autoref{tab:kitti_sequences_lidar} and \autoref{tab:overall_lidar} to Tables 1-4, we can notice that, in general,  both \netname{}  and \netname{}-GT exhibit a lower EPE when steered by LIDAR supervision thanks to the higher depth resolution provided by the sensor. Yet, the  D1-all error tends to be higher. We ascribe this to raw LIDAR measurements featuring a  larger number of outliers compared to SGM and WILD because both the latter pipelines include a filtering step to disregard low-confidence disparities. In particular, 
it is worth observing how raw LIDAR measurements often yield gross errors near depth discontinuities, as illustrated in \autoref{fig:lidar_density}.

\rev{\textbf{Complementing low-density supervision.} As shown in both \autoref{fig:density} and \autoref{fig:lidar_density}, we can notice how different the density of the labels is by changing the source, with WILD, in particular, resulting the method providing sparser supervision. To obtain denser supervision, we might combine WILD with self-supervision \cite{Tonioni_2019_CVPR} or with LIDAR measurements, that often results complementary to it. We show this experiment in the supplementary material.}

\begin{figure}[t]
    \centering
    \begin{tabular}{c}
        \includegraphics[width=0.45\textwidth]{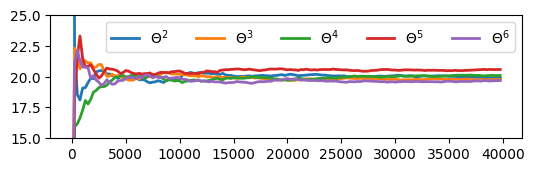}\\
        \includegraphics[width=0.45\textwidth]{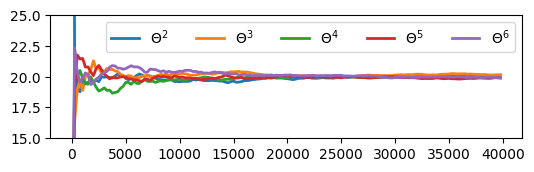}\\
    \end{tabular}
    \caption{\rev{\textbf{Percentage of update steps across \netname{}'s modules over time.} Update frequency (\%) for $\Theta^i, i \in [2,\dots,6]$ \algoname{} (top) and \algoname{}++ with SGM proxy labels (bottom) as function of the processed frames. Experiment dealing with continual adaptation from synthetic pre-training on  \emph{Campus} $\rightarrow$ \emph{City} $\rightarrow$ \emph{Residential}  $\rightarrow$ \emph{Road}.} \label{fig:percentage}}
\end{figure}

\textbf{Distribution of update steps in modular adaptation.} We dig deeper into the behavior of Algo \autoref{algo:mad} in order to study the distribution of the update steps across the modules  of \netname{}.  
\rev{
\autoref{fig:percentage} plots the distribution of update steps (expressed as a percentage of the processed frames) occurring for each module during continual adaptation \textit{in-the-wild} across the four KITTI domains (same setting as in \autoref{tab:overall}): in the top and bottom figures we show the results achieved by running \algoname{} and \algoname{}++ with SGM labels, respectively.
We can notice how, with the original \algoname{} strategy based on self-supervision by the photometric error, most of the steps concern $\Theta_5$, which shows consistently a higher percentage with respect to the other modules, while $\Theta_6$ results the module less frequently updated. Conversely, \algoname{}++ steered by SGM yields a more balanced distribution of updates across the modules, with almost no gap between the percentages observable after 20K frames.  We argue that the more even distribution of the  update steps featured by \algoname{}++ leads to a better approximation of continual adaptation by  back-propagation into the whole network (Algo \autoref{algo:backprop}), which, indeed, updates all modules in each step.}
A round-robin strategy would  allow for uniform sampling of the modules alike, but such a fixed sampling schedule proved to be less effective than \algoname{} \cite{Tonioni_2019_CVPR}. \algoname{}++,  seems to provide a balanced update distribution without being bound to a fixed schedule, which, in fact, turns out beneficial to the continual adaptation process. 
As such, \autoref{fig:percentage} may help explaining why, as observed in \autoref{tab:kitti_sequences} and \autoref{tab:overall}, \algoname{}++ is  more effective than \algoname{} in filling the performance gap between modular adaptation and full adaptation.  

\begin{figure}[t]
    \centering
    \renewcommand{\tabcolsep}{1pt}
    \begin{tabular}{c}
    \includegraphics[width=0.48\textwidth]{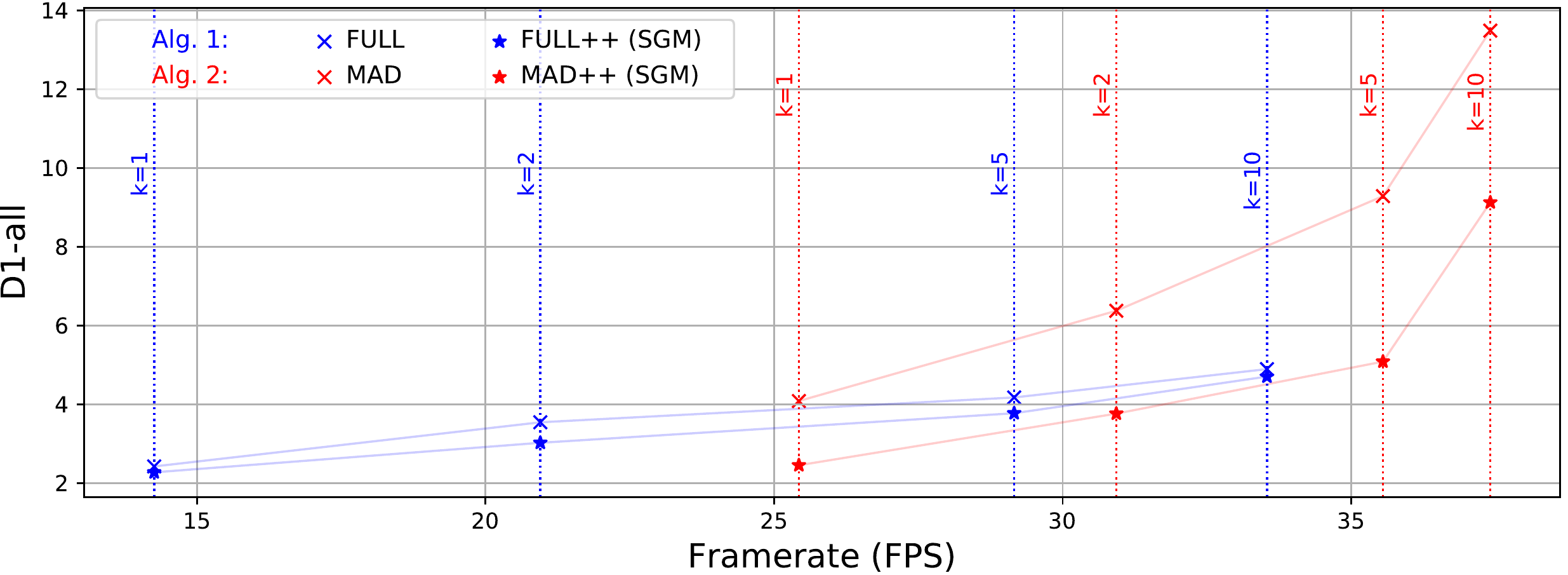}
    \end{tabular}
    \caption{\textbf{Performance vs. speed with different adaptation rates.} 
    FPS vs. D1-all measurements for Algo \autoref{algo:backprop} (blue) and Algo \autoref{algo:mad} (red) when adapting every K=1,2,5,10 frames.  Solid lines represent linear interpolations between measurements. Results dealing with continual adaptation from synthetic pre-training on \emph{Campus} $\rightarrow$ \emph{City} $\rightarrow$ \emph{Residential}  $\rightarrow$ \emph{Road}.
    }
    \label{fig:dilated}
\end{figure}

\textbf{Experiments with different adaptation rates.} To pursue further computational efficiency in continual adaptation, we investigate on updating the model on only a subset of the incoming frames, so as to reduce the computational overhead due to continuously gathering knowledge about the sensed environment. Purposely, we consider a more general adaptation approach, whereby the model is updated every $K$ frames, and evaluate the speed-performance trade-off yielded by Algo \autoref{algo:backprop} and Algo \autoref{algo:mad} while varying $K$.  \autoref{fig:dilated} plots the D1-all error as a function of the frame-rate\footnote{Ideal performance measured on a 1080Ti GPU.} achieved with $K=1,2,5,10$ by considering  self-supervision (FULL and \algoname{}) and proxy supervision (FULL++ and \algoname{}++ with SGM proxies). 

Performing adaptation at every frame (K=1)  allows \netname{} to run at about 14 FPS with FULL and FULL++, while the network can achieve more than 25 FPS by \algoname{} and \algoname{}++. By increasing the interval between subsequent adaptations, we observe a corresponding  increase of both speed and disparity prediction errors for all the considered methods.

\begin{table*}[t]
	\center
	\setlength{\tabcolsep}{7pt}
	\scalebox{0.85}
	{
	\begin{tabular}{|l|ll|ll|ll|ll|}
		\cline{4-9}
		\multicolumn{3}{c}{} & \multicolumn{2}{|c|}{Rainy (1667 frames)} & \multicolumn{2}{c|}{Dusky (1119 frames)} & \multicolumn{2}{c|}{Cloudy (4950 frames)} \\
		\hline
		Starting Model & Adapt. Mode & Proxy src. & D1-all(\%) & EPE & D1-all(\%) & EPE & D1-all(\%) & EPE \\ 
		\hline
		\hline
        \netname{} & No & \xmark & 31.40 & 4.46 & 39.04 & 5.77 & 25.37 & 3.13 \\
        LEAStereo \cite{cheng2020hierarchical} & No & \xmark & 22.14 & 3.90 & 22.99 & 3.45 & 8.64 & 1.61 \\
        \hline
        \hline
		
		\netname{} & FULL & \xmark & 19.64 & 2.71 & 20.40 & 2.48 & 9.54 & 1.57 \\ 
        \netname{} & MAD & \xmark & 26.11 & 3.67 & 33.23 & 5.43 & 16.03 & 2.83 \\ 
        & & & \textcolor{blue}{(+6.47)} & \textcolor{blue}{(+0.97)} & \textcolor{blue}{(+12.84)} & \textcolor{blue}{(+2.96)} & \textcolor{blue}{(+6.49)} & \textcolor{blue}{(+1.25)} \\
		\hline

        \netname{}-K-FULL & FULL & \xmark & 15.75 & 2.46 & 9.21 & 1.43 & 8.09 & 1.48 \\ 
        \netname{}-K-MAD & MAD & \xmark & 12.22 & 1.82 & 9.81 & 1.43 & 7.32 & 1.41 \\ 
         & & & \textcolor{blue}{(-3.53)} & \textcolor{blue}{(-0.63)} & \textcolor{blue}{(+0.60)} & \textcolor{blue}{(0.00)} & \textcolor{blue}{(-0.77)} & \textcolor{blue}{(-0.06)} \\ 
        \hline\hline		
		
		\netname{} & FULL++ & SGM \cite{hirschmuller2005accurate} & 17.28 & 2.62 & 12.86 & 1.73 & 6.63 & 1.33 \\ 
        \netname{} & MAD++ & SGM \cite{hirschmuller2005accurate} & 17.70 & 2.48 & 13.61 & 1.84 & 7.76 & 1.45 \\ 
        & & & \textcolor{blue}{(+0.42)} & \textcolor{blue}{(-0.15)} & \textcolor{blue}{(+0.76)} & \textcolor{blue}{(+0.11)} & \textcolor{blue}{(+1.13)} & \textcolor{blue}{(+0.12)} \\
		\hline

        \netname{}-K-FULL++ & FULL++ & SGM \cite{hirschmuller2005accurate} & 12.97 & 2.50 & 6.99 & 1.47 & 6.61 & 1.66 \\ 
        \netname{}-K-MAD++ & MAD++ & SGM \cite{hirschmuller2005accurate} & 12.65 & 2.32 & 5.93 & 1.40 & 6.26 & 1.72 \\ 
         & & & \textcolor{blue}{(-0.33)} & \textcolor{blue}{(-0.18)} & \textcolor{blue}{(-1.06)} & \textcolor{blue}{(-0.07)} & \textcolor{blue}{(-0.35)} & \textcolor{blue}{(+0.06)} \\
        \hline        
		SGM \cite{hirschmuller2005accurate} & \xmark & \xmark & 17.88 & 6.55 & 11.83 & 2.60 &  7.31 & 2.33 \\
		&   &   & (4.59) & (1.25) & (4.49) & (1.01) & (2.10) & (0.86) \\
		&   &   & \multicolumn{2}{c|}{(66.90\% density)} & \multicolumn{2}{c|}{(77.35\% density)} & \multicolumn{2}{c|}{(79.53\% density)} \\
        \hline\hline
		
		\netname{} & FULL++ & WILD \cite{Tosi_2017_BMVC} & 17.93 & 2.60 & 19.04 & 2.32 & 8.31 & 1.49 \\ 
        \netname{} & MAD++ & WILD \cite{Tosi_2017_BMVC} & 17.71 & 2.40 & 19.24 & 2.48 & 8.38 & 1.52 \\ 
        & & & \textcolor{blue}{(-0.22)} & \textcolor{blue}{(-0.20)} & \textcolor{blue}{(+0.20)} & \textcolor{blue}{(+0.16)} & \textcolor{blue}{(+0.07)} & \textcolor{blue}{(+0.03)} \\
        \hline

        \netname{}-K-FULL++ & FULL++ & WILD \cite{Tosi_2017_BMVC} & 14.01 & 2.24 & 7.94 & 1.30 & 6.21 & 1.30 \\ 
        \netname{}-K-MAD++ & MAD++ & WILD \cite{Tosi_2017_BMVC} & 13.70 & 2.09 &
7.50 & 1.27 & 5.96 & 1.29 \\ 
         & & & \textcolor{blue}{(-0.31)} & \textcolor{blue}{(-0.15)} & \textcolor{blue}{(-0.44)} & \textcolor{blue}{(-0.03)} & \textcolor{blue}{(-0.25)} & \textcolor{blue}{(-0.01)} \\

        \hline         
        WILD \cite{Tosi_2017_BMVC} & \xmark & \xmark & 36.55 & 17.85 & 33.55 & 13.83 &  22.89 & 9.77 \\
		&   &   & (2.45) & (1.03) & (1.92) & (0.87) & (1.13) & (0.83) \\
		&   &   & \multicolumn{2}{c|}{(23.37\% density)} & \multicolumn{2}{c|}{(25.12\% density)} & \multicolumn{2}{c|}{(28.21\% density)} \\
		\hline
		\multicolumn{9}{c}{ } \\
	\end{tabular}
	}
	\vspace{1pt}
	\caption{\textbf{Online adaptation on DrivingStereo.} Results on the  \emph{Rainy}, \emph{Dusky} and \emph{Cloudy} sequences. \emph{-K} denotes prior adaption on \kitti{}  (\emph{Campus} $\rightarrow$ \emph{City} $\rightarrow$ \emph{Residential}  $\rightarrow $ \emph{Road}) before further adaptation on DrivingStereo.}		
	\label{tab:drivingstereo}
\end{table*}

We can appreciate the better performance-speed trade-off provided by \algoname{}++ by fixing a target frame-rate and comparing the accuracy yielded by the different methods, or vice-versa. For instance, considering 30 FPS as the target speed requirement, we can notice that this is met by Algo \autoref{algo:mad} and Algo \autoref{algo:backprop} with K=2 and k=10, respectively. In these settings, \algoname{}++ achieves the lowest D1-all error, with FULL and FULL++ running faster but yielding higher errors and the original \algoname{} formulation resulting both slower and less accurate than both variants of Algo \autoref{algo:backprop}.
 Should the application demand  a  lower real-time requirement, \ie{} 25 FPS,  \algoname{}++ would turn out again the most accurate method to meet the given target speed. 
On the other hand, we may compare methods  achieving close D1-all scores, \eg{} \algoname{}++ and FULL++, both with K=1, and highlight how \algoname{}++ can run much faster with equivalent accuracy. Likewise, by considering Algo \autoref{algo:backprop} with K=5 and Algo \autoref{algo:mad} with K=2, we can observe how \algoname{} would turn out significantly less accurate than FULL and FULL++, whilst \algoname{}++ can provide equivalent accuracy and faster speed. 

Thus,  we are lead to conclude that \algoname{}++ achieves a more favourable performance-speed  trade-off with respect to the other continual adaptation methods, as also suggested in \autoref{fig:dilated} by the trend of the interpolating curves (the lower the better).

\subsection{Evaluation on DrivingStereo} \label{sec:drivingstereo}

Here we move to a different dataset dealing with autonomous driving scenarios in order to assess the performance of continual adaptation in diverse weather conditions. Purposely, we select three sequences from the DrivingStereo dataset \cite{Yang_2019_CVPR}:  \textit{Rainy} and \textit{Dusky} are short videos featuring less than 2k frames, while \textit{Cloudy} counts about 5k frames. Moreover, \textit{Rainy} depicts imagery acquired in presence of rain and wet road surface, making it particularly hard for traditional stereo algorithms too.
These three sequences have been selected  to study: 1) how continual adaptation performs in presence of challenging weather conditions (\eg{}, rain),  2) how prior continual adaptation does affect performance when facing a new environment. Short sequences allow to better investigate on the latter issue, as  long ones  would hide the effects of prior adaptation on the initial frames due to performance figures being  averaged  across all frames.
\rev{As for experiments concerning prior continual adaptation, we use the \netname{} instance trained on synthetic images and adapted on \emph{Campus} $\rightarrow$ \emph{City} $\rightarrow$ \emph{Residential}  $\rightarrow $ \emph{Road}, dubbed as \netname{}-K-A, with A being the specific adaptation mode adopted.}

\textbf{Keep adapting!} \autoref{tab:drivingstereo} collects the experimental results on DrivingStereo. In the first row we report the baseline performance achieved by running \netname{} after pre-training on synthetic imagery and without any further adaptation, \rev{followed by the results achieved by the current top-1 method on the KITTI online benchmark pre-trained on the same synthetic data, \ie LEAStereo \cite{cheng2020hierarchical}, that highlights how current state-of-the-art methods are prone to the domain-shift issue.} The three successive sub-tables, \ie{} rows 2-7, 8-16 and 17-25, collect results dealing with continual adaptation realized through self-supervision by the photometric error loss, proxy supervision by SGM and proxy supervision by WILD, respectively. For the two sources of proxy labels we also show the errors yielded by the pipeline providing the supervision, both before and after the outlier removal step (rows 14-15 and 23-24), alongside the resulting label density (rows 16 and 25). We do not consider supervision by LIDAR as such measurements are not provided in DrivingStereo.  

Considering self-supervision, we first adapt on a sequence by FULL and \algoname{} (rows 2 and 3). As observed on KITTI, both strategies are effective, though a large margin between the two exists. In rows 5 and 6 we report the results achieved by  FULL (\netname{}-K-FULL) and \algoname{} (\netname-K-\algoname{}) with model instances obtained through prior continual adaptation on KITTI (\ie, the models saved after the experiments in \autoref{tab:overall}). Hence, by keep adapting to the current domain we achieve much better results, as highlighted by the comparison between rows 5 and 2 as well as 6 and 3. 
In particular, prior adaptation is particularly effective in the short-term, as shown by the \textit{Dusky} sequence where the error rate is roughly halved with FULL and brought down by about 70\% with \algoname{}, while the benefit tends to be  smaller in longer sequences, such as \textit{Cloudy}, in particular with FULL. These experimental findings show  that continual adaptation realized through both FULL and \algoname{} is always beneficial. 
Interestingly, we point out how prior adaptation turns out much more effective with MAD, allowing it to even outperform FULL in several sequences (as shown by row 7).

Moving to the experiments dealing with proxy supervision by SGM, we can notice, in general, a substantial performance improvement. In particular, when adapting \netname{} starting from 
pre-training on synthetic imagery, SGM proxies (rows 8 and 9) consistently outperform self-supervision by the
photometric error loss (rows 2 and 3) and allow for breaking down the  margin between Algo \autoref{algo:backprop} and Algo \autoref{algo:mad}  (row 10 vs row 4). 
By keep adapting from KITTI (rows 11 and 12), results get much better on all sequences and SGM neatly outperforms   adoption by the photometric loss in the same training protocol (rows 5 and 6). In this case, moreover, \algoname{}++ outperforms  FULL++ on all sequences (as highlighted in row 13).

Finally, by analyzing the experimental results obtained by WILD proxies, we can observe a trend similar to that already discussed for SGM. In particular, keep adapting on DrivingStereo following prior adaptation on KITTI is highly beneficial and  \algoname{}++  turns out more effective than FULL++ in this setting (row 22).

\textbf{Proxy labels comparison.} We use DrivingStereo sequences also to further evaluate proxy labels. Rows 14-16 and 23-25 report  the accuracy and density of the labels produced by SGM and WILD. In particular,  we show first the error rates achieved without filtering out the outliers, then, in brackets, those computed only on the final labels used to provide the supervision and finally  the density of the filtered labels. We can notice that the SGM pipeline extracts much more labels at the cost of a lower accuracy. This confirms the finding already discussed in 
\autoref{sec:kitti}
about the different traits of the two pipelines deployed to attain proxy supervision.

Eventually, we point out that \netname{} models adapted by high-confidence (\ie filtered)  SGM and WILD proxies tend to consistently outperform the dense (\ie{}, without filtering) pipeline providing the supervision, more often than not by a large margin. Indeed, only with \emph{Dusky} (the shortest sequence) the SGM pipeline (row 14) slightly outperforms \netname{} adapted from synthetic pre-training (rows 8 and 9), though when keep adapting the model following prior adaptation on KITTI the latter turns out about 40-50\% more accurate (rows 11 and 12). 

\rev{However, the accuracy of the filtered points remains higher, yet much sparser and thus less amenable for practical applications. Although approaches to restore density exists \cite{aleotti2020reversing} and might represent an alternative to our framework, in the supplementary material we show that, in practise, using the sparse proxies to adapt \netname{} often leads to better results.}

\begin{table}[t]
	\setlength{\tabcolsep}{9pt}
	\center
    \scalebox{0.82}
    {
			\begin{tabular}{|l|ll|l|}
				\hline
				Starting Model & Adapt. Mode & Proxy src. & Photometric Error \\ 
				\hline
				\netname{} & No & \xmark & 0.124 \\
				\hline\hline
				\netname{} & FULL & \xmark & 0.084 \\
				\netname{} & MAD & \xmark & 0.094 \\
				& & & \textcolor{blue}{(+0.010)} \\
				\hline
				\netname{}-K-FULL & FULL & \xmark & 0.080 \\
				\netname{}-K-MAD & MAD & \xmark & 0.083 \\
				& & & \textcolor{blue}{(+0.003)} \\
                \hline\hline
				\netname{} & FULL++ & SGM \cite{hirschmuller2005accurate} & 0.086 \\
				\netname{} & MAD++ & SGM \cite{hirschmuller2005accurate} & 0.088 \\
				& & & \textcolor{blue}{(+0.002)} \\
				\hline
				\netname{}-K-FULL++ & FULL++ & SGM \cite{hirschmuller2005accurate} & 0.082 \\
				\netname{}-K-MAD++ & MAD++ & SGM \cite{hirschmuller2005accurate} & 0.082 \\
				& & & \textcolor{blue}{(0.000)} \\
                \hline\hline
				\netname{} & FULL++ & WILD \cite{Tosi_2017_BMVC} & 0.087 \\
				\netname{} & MAD++ & WILD \cite{Tosi_2017_BMVC} & 0.088 \\
				& & & \textcolor{blue}{(+0.001)} \\
				\hline
				\netname{}-K-FULL++ & FULL++ & WILD \cite{Tosi_2017_BMVC} & 0.082 \\
				\netname{}-K-MAD++ & MAD++ & WILD \cite{Tosi_2017_BMVC} & 0.082 \\
				& & & \textcolor{blue}{(0.000)} \\				
				\hline
			\end{tabular}
			}
	\vspace{1pt}
		\caption{\textbf{Online adaptation on WeanHall.} Results on the  \emph{Rainy}, \emph{Dusky} and \emph{Cloudy} sequences. \emph{-K} denotes prior adaption on \kitti{}  (\emph{Campus} $\rightarrow$ \emph{City} $\rightarrow$ \emph{Residential}  $\rightarrow $ \emph{Road}) before further adaptation on WeanHall.}		
	\label{tab:weanhall}
\end{table}

\subsection{Evaluation on WeanHall}

Finally, we make a further step into evaluating the effectiveness of continual adaptation across very different domains by moving to an indoor environment, like that featured by the WeanHall dataset. As previously pointed out, given the absence of ground-truth labels for this set of images, we measure the photometric error according to \autoref{eq:photo}.

\autoref{tab:weanhall} collects the outcome of the experiments carried out on WeanHall following the same protocol adopted in \autoref{sec:drivingstereo} for DrivingStereo. Although the margins in terms of photometric error are smaller compared to the  D1-all and EPE metrics, we observe findings consistent to previous experiments. Indeed, continual adaptation yields a significant improvement with respect to the baseline, with the adoption of proxy labels rather than self-supervision shrinking the gap between Algo \autoref{algo:backprop} and Algo \autoref{algo:mad}, and \algoname{}++ always outperforming \algoname{}. 
More importantly, starting from models previously adapted on KITTI always lead to better results, despite the successive  adaptation being conducted in a totally different environment (\ie indoor vs. outdoor). 

\begin{table}[t]
	\center
	\setlength{\tabcolsep}{2pt}
	\scalebox{0.85}
	{
	\begin{tabular}{|ll|ll|ll|ll|}
		\cline{3-8}
		\multicolumn{2}{c}{ } & \multicolumn{2}{|c|}{Before WeanHall} & \multicolumn{2}{c|}{After WeanHall} & \multicolumn{2}{c|}{After WeanHall} \\
        \multicolumn{2}{c}{ } & \multicolumn{2}{|c|}{ } & \multicolumn{2}{c|}{ } & \multicolumn{2}{c|}{(adapting)} \\	
        \hline
		Adapt. Mode & Proxy src. & D1-all(\%) & EPE & D1-all(\%) & EPE & D1-all(\%) & EPE \\
		\hline
		\hline
		FULL & \xmark & 2.43 & 0.95 & 2.03 & 0.93 & 1.49 & 0.87\\
        MAD & \xmark & 4.09 & 1.19& 1.34 & 0.77 & 2.43 & 0.95 \\
        & & \textcolor{blue}{(+1.66)} & \textcolor{blue}{(+0.24)} & \textcolor{blue}{(-0.69)} & \textcolor{blue}{(-0.16)} & \textcolor{blue}{(+0.94)} & \textcolor{blue}{(+0.08)}\\
        \hline
		FULL++ & SGM & 2.28 & 0.95 & 2.45 & 0.94 & 1.51 & 0.88 \\
        MAD++ & SGM & 2.46 & 0.98 & 2.52 & 0.95 & 1.94 & 0.91\\
        & & \textcolor{blue}{(+0.18)} & \textcolor{blue}{(+0.03)} & \textcolor{blue}{(+0.07)} & \textcolor{blue}{(+0.01)} & \textcolor{blue}{(+0.43)} & \textcolor{blue}{(+0.03)}\\
        \hline
	\end{tabular}
	}
	\vspace{3pt}
	\caption{\rev{\textbf{Experimental results concerning catastrophic forgetting.} Results on the \emph{City}$\rightarrow{}$\emph{Residential}$\rightarrow{}$\emph{Campus}$\rightarrow{}$\emph{Road} sequences from \kitti{}. From left to right, results from first adaptation on KITTI (\autoref{tab:overall}) and from a second run on KITTI after having adapted on WeanHall, respectively without and with adaptation enabled during the second run. Starting model: \netname{}.}	}	
	\label{tab:forgetting}
\end{table}

\rev{
\subsection{Catastrophic forgetting experiment}

Finally, to highlight how our framework is not affected by catastrophic forgetting during continuous adaptation across very different domains, we perform a second adaptation run on the entire KITTI raw dataset, after having adapted the model first on KITTI and then on WeanHall. Should  catastrophic forgetting occur, the accuracy of \netname{} during this second run would dramatically drop. \autoref{tab:forgetting} collects the outcome of this experiment, showing the results achieved by \netname{} with different adaptation strategies. In particular, we report, from left to right, the results concerning  1) the first run over KITTI (i.e., those shown in \autoref{tab:overall}, 2) the second run over KITTI after having adaptation on WeanHall, without further adaption on KITTI and 3) the second run over KITTI while keeping adapting. For FULL++/MAD++, we show the results achieved when using SGM proxy labels.
We can notice how, even without adapting a second time to KITTI, the accuracy is comparable to that achieved during the first run, highlighting that catastrophic forgetting has not occurred, \ie upon adaptation on WeanHall the network has not forgotten how to predict disparities on KITTI. Indeed, with FULL/MAD the results are improved, while with FULL++/MAD++ we observe a negligible drop. 
By turning  adaptation on in the second run on KITTI we observe how performance can further improve, except for MAD.}

\begin{figure}[t]
    \centering
    \renewcommand{\tabcolsep}{1pt}
    \begin{tabular}{cccc}
        \multirow{2}{*}{\rotatebox{90}{Frame 0}} & \begin{overpic}[width=0.155\textwidth]{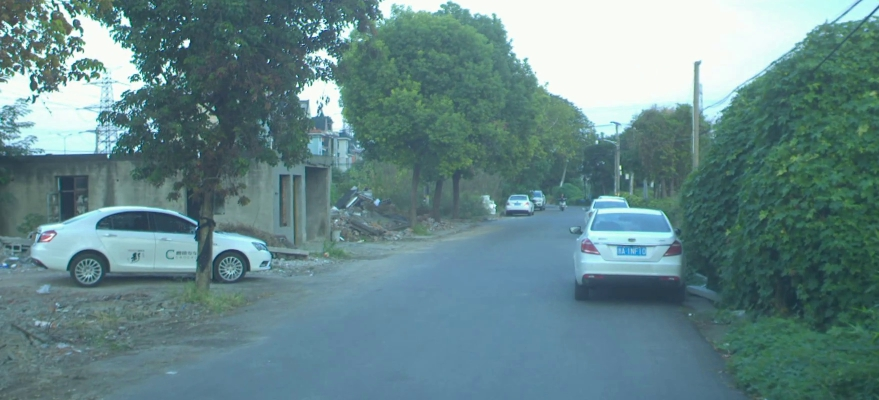} 
        \put (0,36) {\colorbox{gray}{$\scriptsize\displaystyle\textcolor{white}{\textbf{LEFT}}$}}
        \end{overpic} &  
        \begin{overpic}[width=0.155\textwidth]{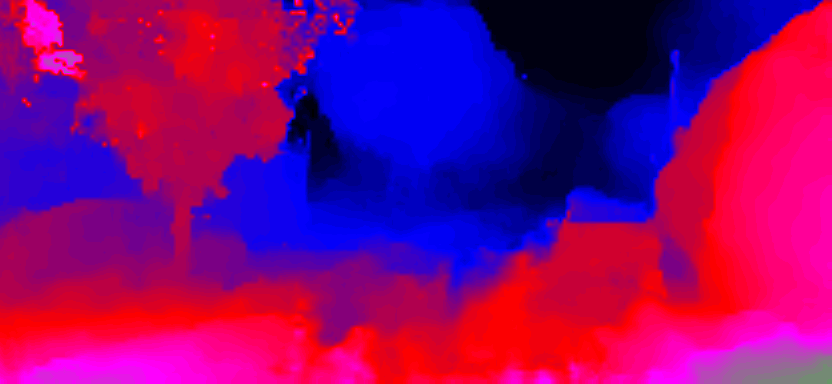}
        \put (0,36) {\colorbox{gray}{$\scriptsize\displaystyle\textcolor{white}{\textbf{MAD}}$}}
        \end{overpic} &
        \begin{overpic}[width=0.155\textwidth]{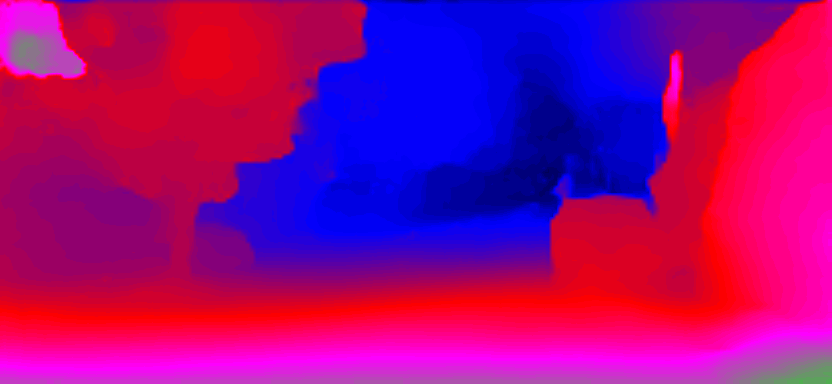}
        \put (0,36) {\colorbox{gray}{$\scriptsize\displaystyle\textcolor{white}{\textbf{K-MAD}}$}}
        \end{overpic} \\
        & \begin{overpic}[width=0.155\textwidth]{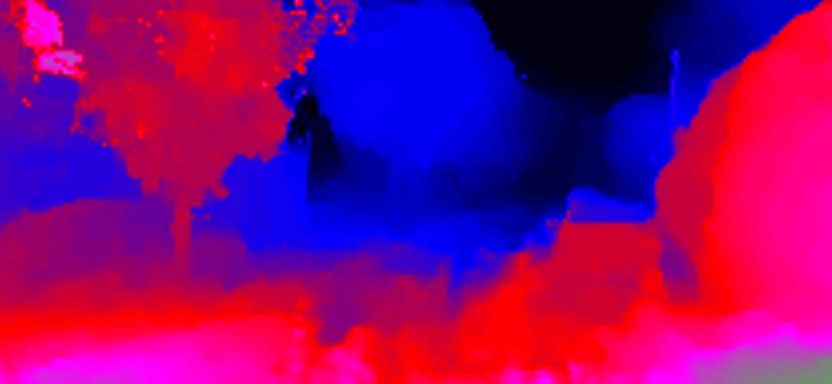} 
        \put (0,36) {\colorbox{gray}{$\scriptsize\displaystyle\textcolor{white}{\textbf{NONE}}$}}
        \end{overpic} &  
        \begin{overpic}[width=0.155\textwidth]{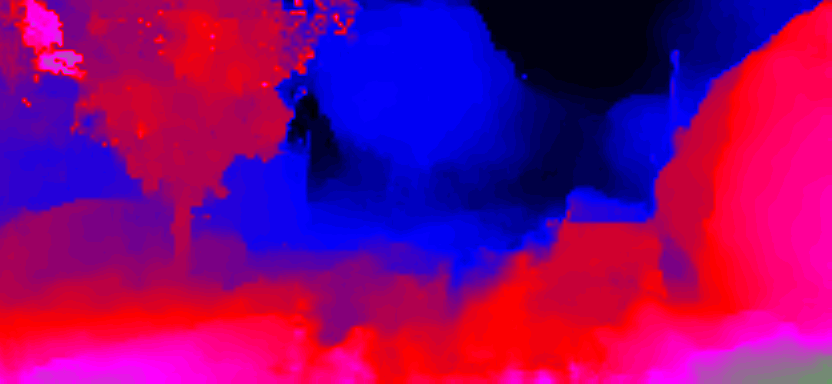}
        \put (0,36) {\colorbox{gray}{$\scriptsize\displaystyle\textcolor{white}{\textbf{MAD++}}$}}
        \end{overpic} &
        \begin{overpic}[width=0.155\textwidth]{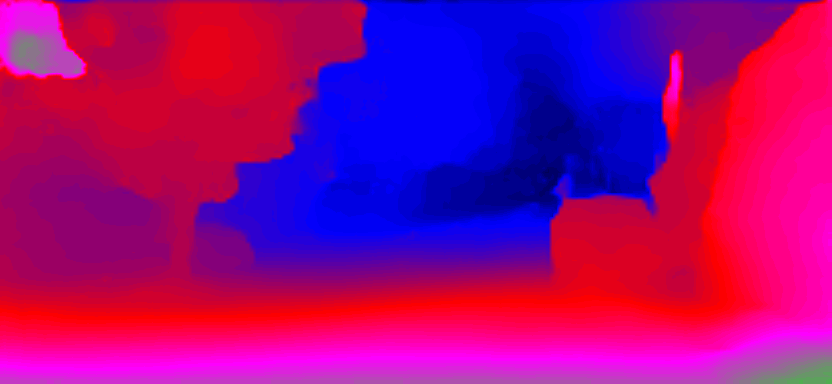}
        \put (0,36) {\colorbox{gray}{$\scriptsize\displaystyle\textcolor{white}{\textbf{K-MAD++}}$}}
        \end{overpic} \\
        \\
        \multirow{2}{*}{\rotatebox{90}{Frame 300}} & \begin{overpic}[width=0.155\textwidth]{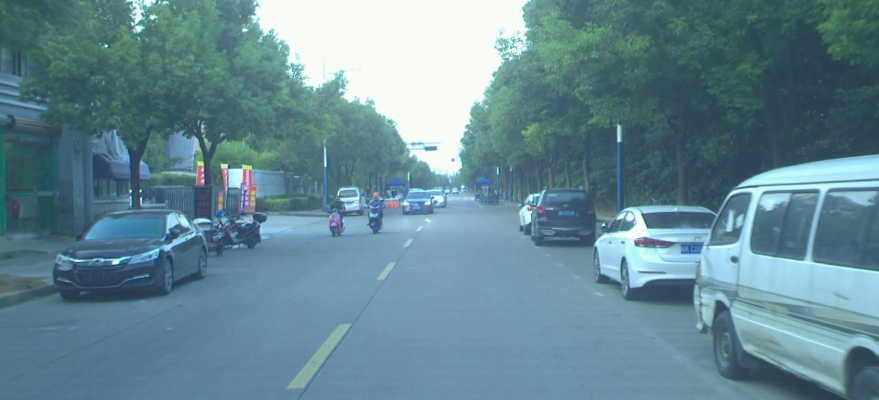} 
        \put (0,36) {\colorbox{gray}{$\scriptsize\displaystyle\textcolor{white}{\textbf{LEFT}}$}}
        \end{overpic} &  
        \begin{overpic}[width=0.155\textwidth]{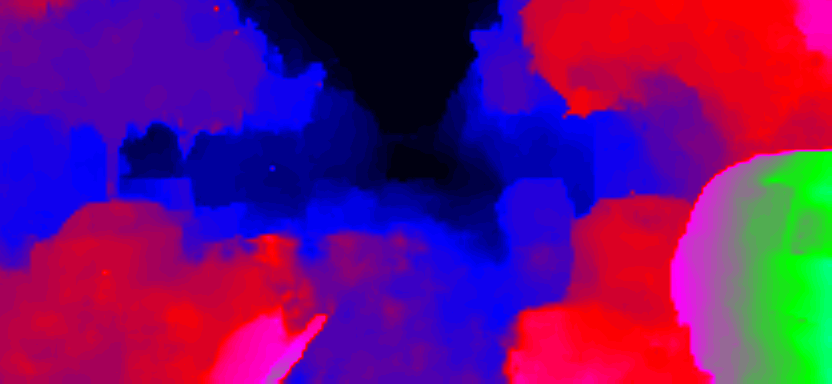}
        \put (0,36) {\colorbox{gray}{$\scriptsize\displaystyle\textcolor{white}{\textbf{MAD}}$}}
        \end{overpic} &
        \begin{overpic}[width=0.155\textwidth]{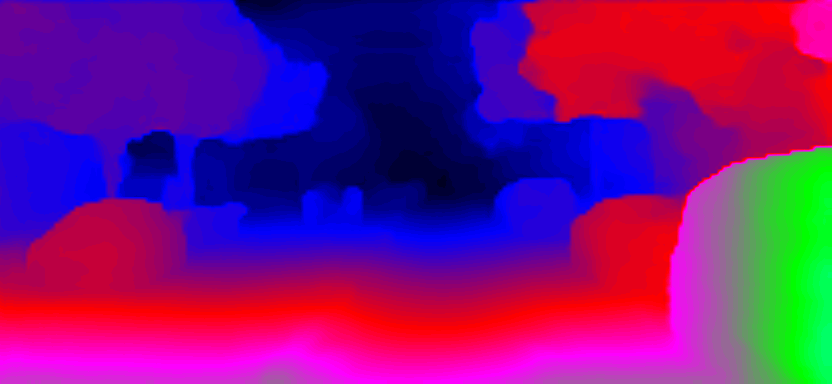}
        \put (0,36) {\colorbox{gray}{$\scriptsize\displaystyle\textcolor{white}{\textbf{K-MAD}}$}}
        \end{overpic} \\
        & \begin{overpic}[width=0.155\textwidth]{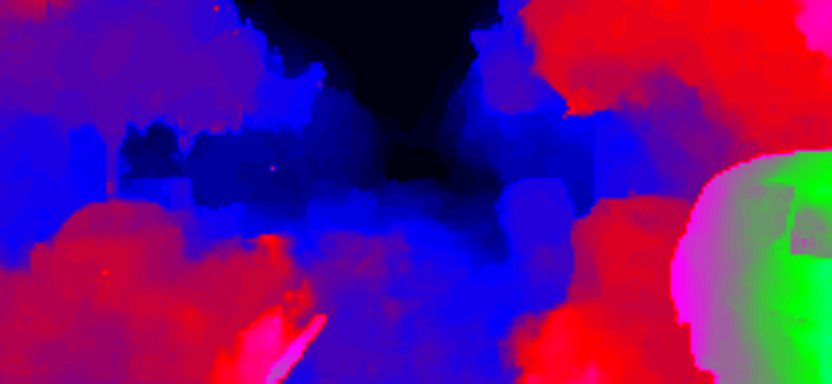} 
        \put (0,36) {\colorbox{gray}{$\scriptsize\displaystyle\textcolor{white}{\textbf{NONE}}$}}
        \end{overpic} &  
        \begin{overpic}[width=0.155\textwidth]{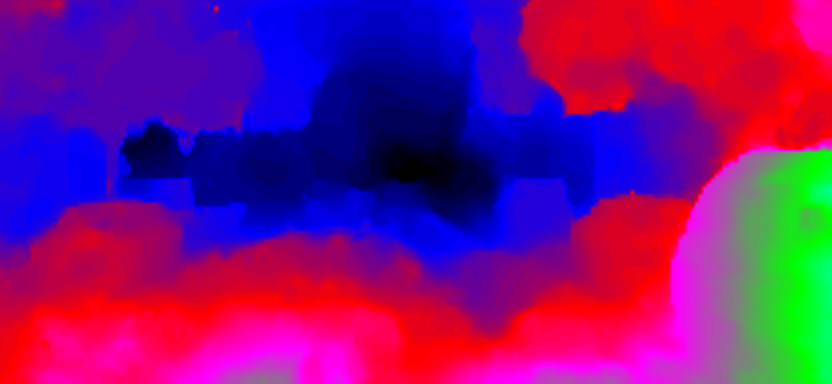}
        \put (0,36) {\colorbox{gray}{$\scriptsize\displaystyle\textcolor{white}{\textbf{MAD++}}$}}
        \end{overpic} &
        \begin{overpic}[width=0.155\textwidth]{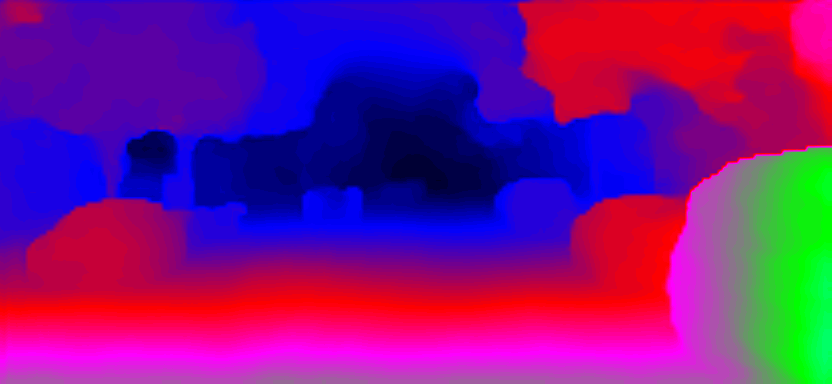}
        \put (0,36) {\colorbox{gray}{$\scriptsize\displaystyle\textcolor{white}{\textbf{K-MAD++}}$}}
        \end{overpic} \\
        \\
        \multirow{2}{*}{\rotatebox{90}{Frame 900}} & \begin{overpic}[width=0.155\textwidth]{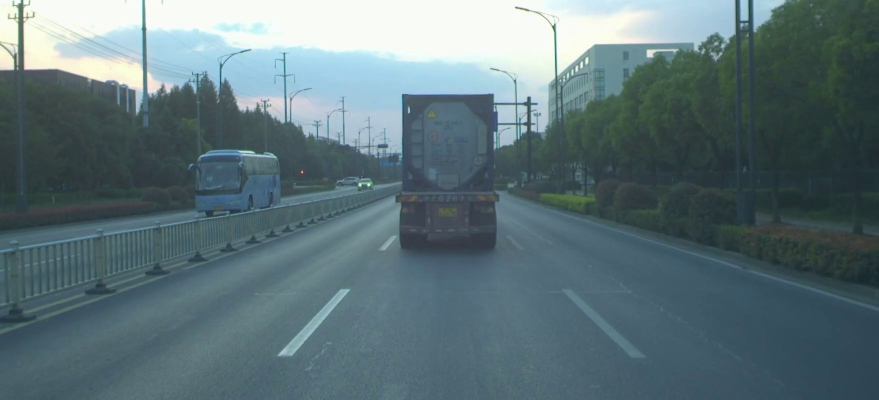} 
        \put (0,36) {\colorbox{gray}{$\scriptsize\displaystyle\textcolor{white}{\textbf{LEFT}}$}}
        \end{overpic} &  
        \begin{overpic}[width=0.155\textwidth]{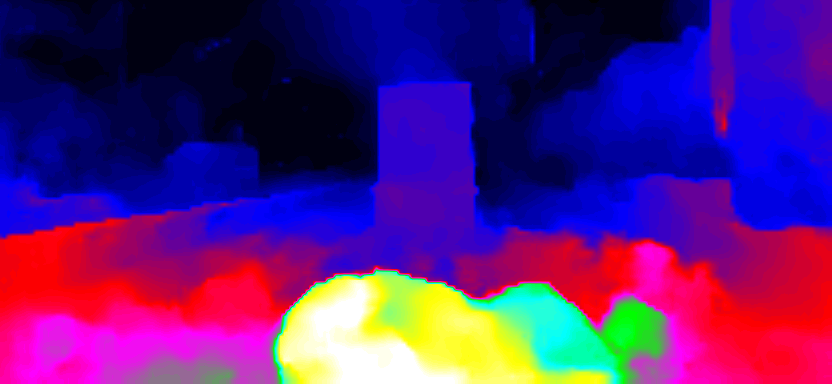}
        \put (0,36) {\colorbox{gray}{$\scriptsize\displaystyle\textcolor{white}{\textbf{MAD}}$}}
        \end{overpic} &
        \begin{overpic}[width=0.155\textwidth]{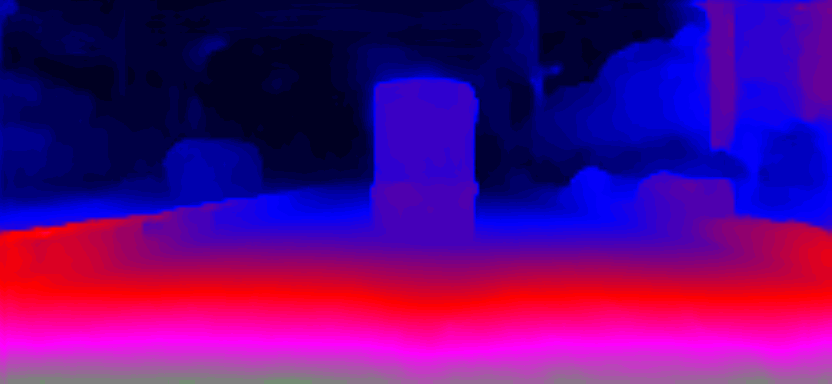}
        \put (0,36) {\colorbox{gray}{$\scriptsize\displaystyle\textcolor{white}{\textbf{K-MAD}}$}}
        \end{overpic} \\
        & \begin{overpic}[width=0.155\textwidth]{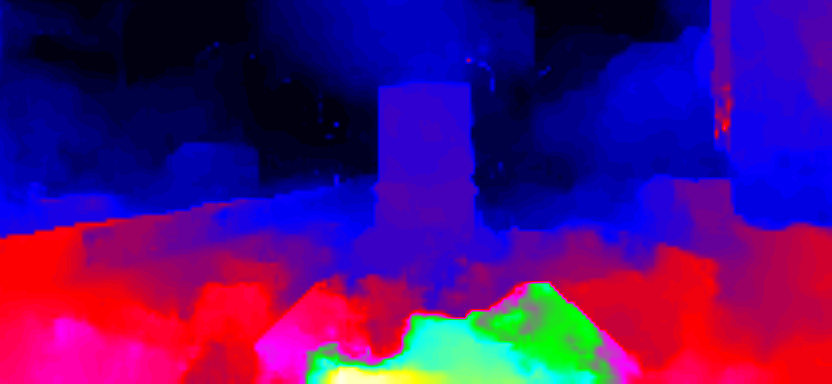} 
        \put (0,36) {\colorbox{gray}{$\scriptsize\displaystyle\textcolor{white}{\textbf{NONE}}$}}
        \end{overpic} &  
        \begin{overpic}[width=0.155\textwidth]{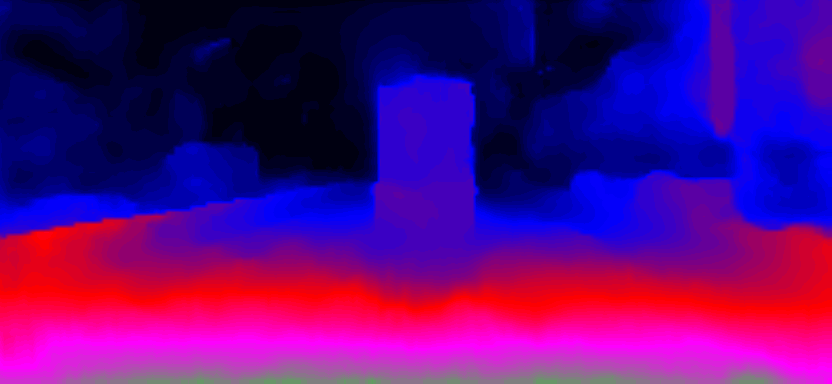}
        \put (0,36) {\colorbox{gray}{$\scriptsize\displaystyle\textcolor{white}{\textbf{MAD++}}$}}
        \end{overpic} &
        \begin{overpic}[width=0.155\textwidth]{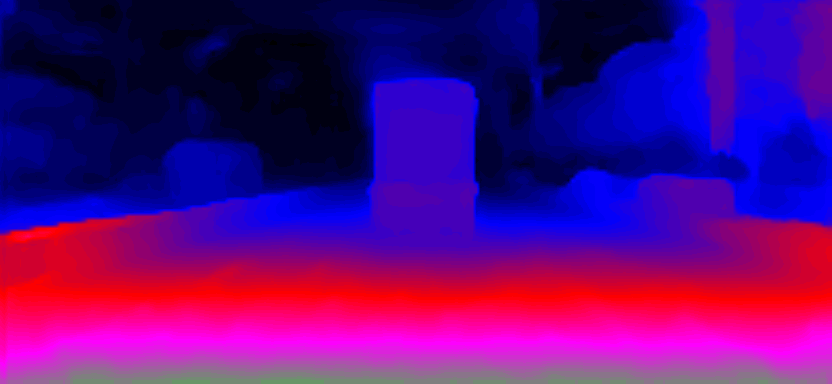}
        \put (0,36) {\colorbox{gray}{$\scriptsize\displaystyle\textcolor{white}{\textbf{K-MAD++}}$}}
        \end{overpic} \\           
    \end{tabular}
    \caption{\rev{\textbf{Qualitative results for different continual adaptation strategies.} We show, across time, the reference image of a stereo pair from DrivingStereo \textit{Dusky} sequence and different adaptation strategies among those reported in \autoref{tab:drivingstereo} (MAD++ uses SGM labels).}}
    \label{fig:qualitatives_driving}
\end{figure}

\begin{figure}[t]
    \centering
    \renewcommand{\tabcolsep}{1pt}
    \begin{tabular}{cccc}
        \multirow{2}{*}{\rotatebox{90}{Frame 0}} & \begin{overpic}[width=0.12\textwidth]{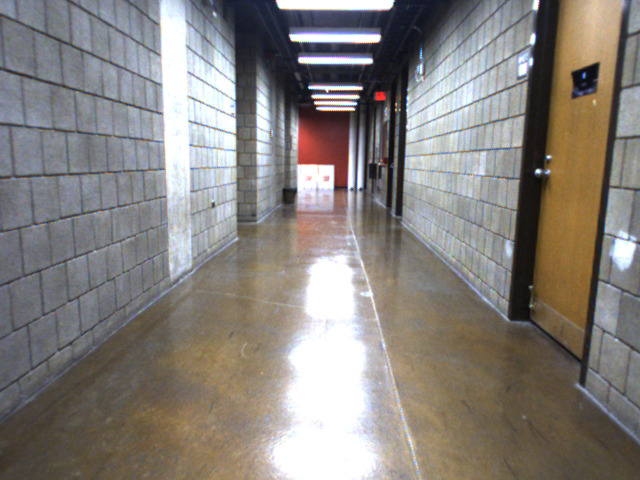} 
        \put (0,60) {\colorbox{gray}{$\scriptsize\displaystyle\textcolor{white}{\textbf{LEFT}}$}}
        \end{overpic} &  
        \begin{overpic}[width=0.12\textwidth]{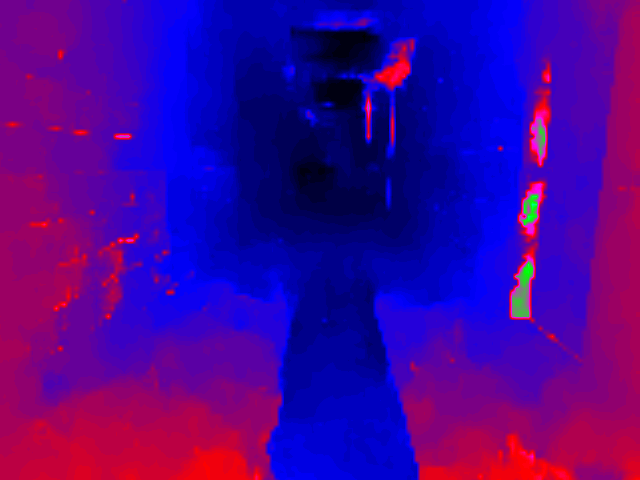}
        \put (0,60) {\colorbox{gray}{$\scriptsize\displaystyle\textcolor{white}{\textbf{MAD}}$}}
        \end{overpic} &
        \begin{overpic}[width=0.12\textwidth]{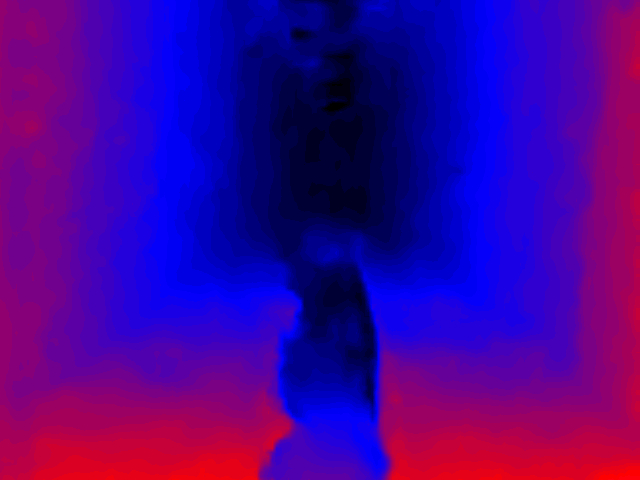}
        \put (0,60) {\colorbox{gray}{$\scriptsize\displaystyle\textcolor{white}{\textbf{K-MAD}}$}}
        \end{overpic} \\
        & \begin{overpic}[width=0.12\textwidth]{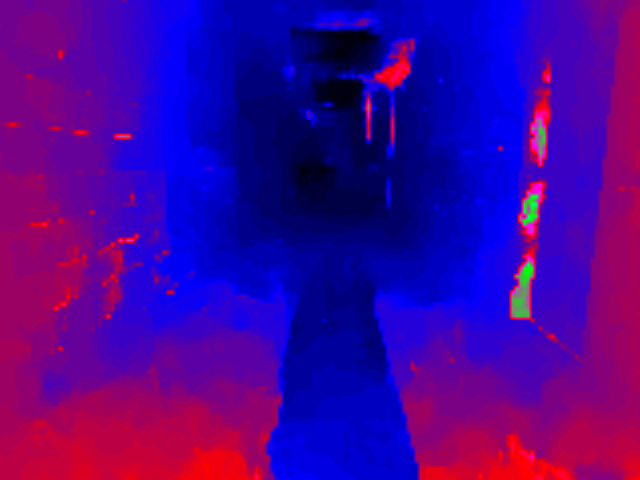} 
        \put (0,60) {\colorbox{gray}{$\scriptsize\displaystyle\textcolor{white}{\textbf{NONE}}$}}
        \end{overpic} &  
        \begin{overpic}[width=0.12\textwidth]{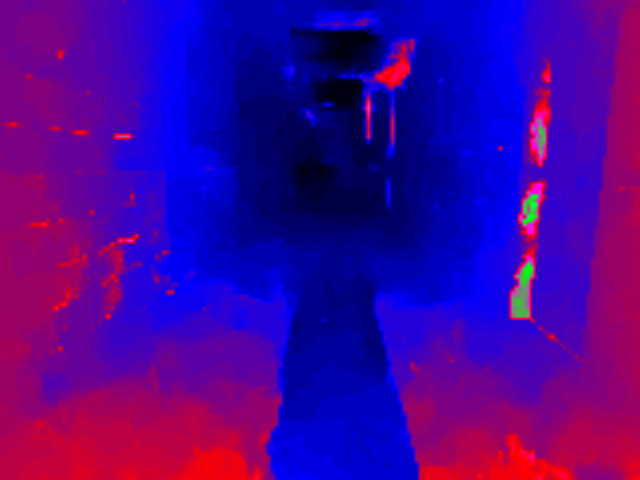}
        \put (0,60) {\colorbox{gray}{$\scriptsize\displaystyle\textcolor{white}{\textbf{MAD++}}$}}
        \end{overpic} &
        \begin{overpic}[width=0.12\textwidth]{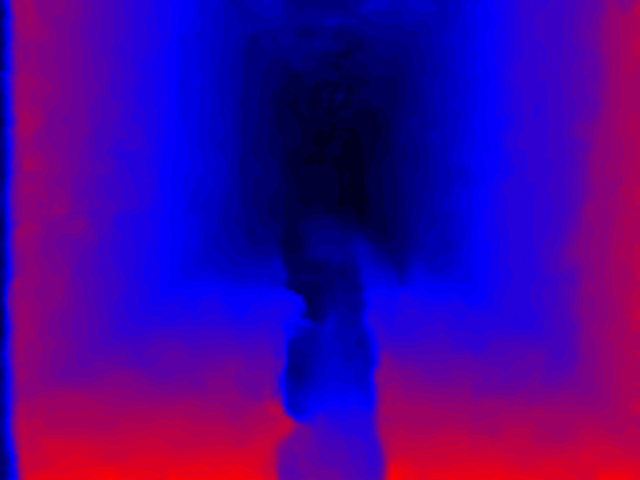}
        \put (0,60) {\colorbox{gray}{$\scriptsize\displaystyle\textcolor{white}{\textbf{K-MAD++}}$}}
        \end{overpic} \\
        \\
        \multirow{2}{*}{\rotatebox{90}{Frame 300}} & \begin{overpic}[width=0.12\textwidth]{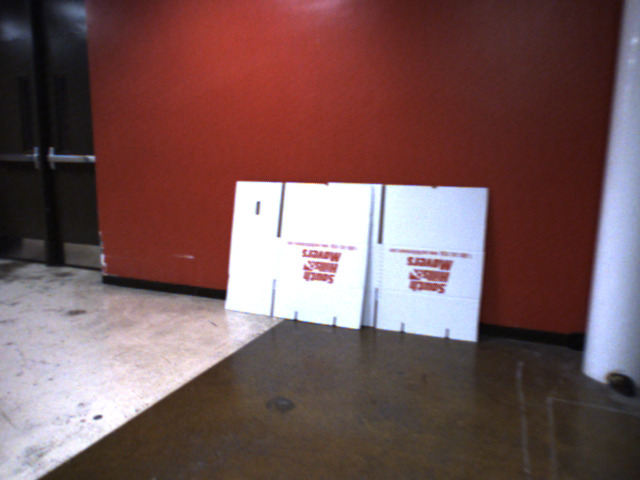} 
        \put (0,60) {\colorbox{gray}{$\scriptsize\displaystyle\textcolor{white}{\textbf{LEFT}}$}}
        \end{overpic} &  
        \begin{overpic}[width=0.12\textwidth]{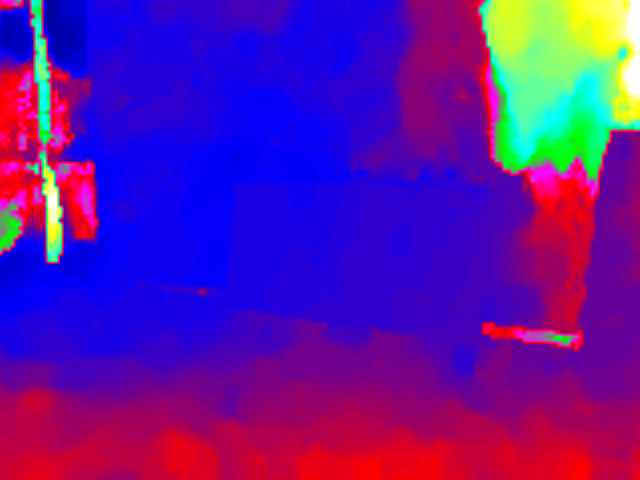}
        \put (0,60) {\colorbox{gray}{$\scriptsize\displaystyle\textcolor{white}{\textbf{MAD}}$}}
        \end{overpic} &
        \begin{overpic}[width=0.12\textwidth]{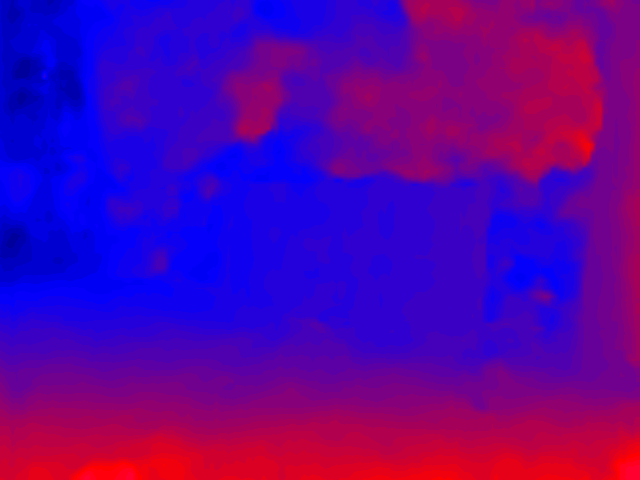}
        \put (0,60) {\colorbox{gray}{$\scriptsize\displaystyle\textcolor{white}{\textbf{K-MAD}}$}}
        \end{overpic} \\
        & \begin{overpic}[width=0.12\textwidth]{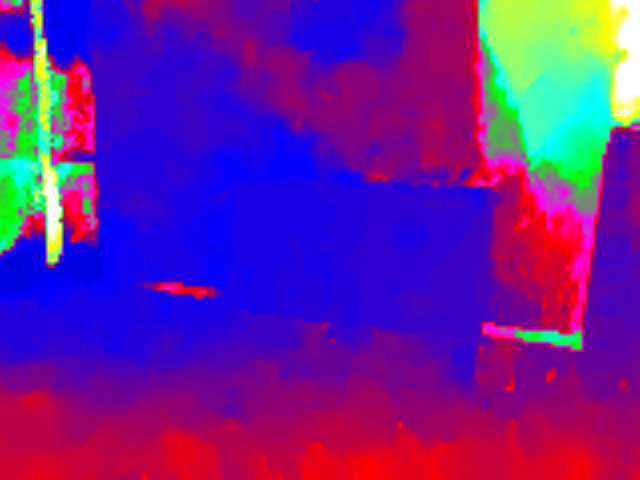} 
        \put (0,60) {\colorbox{gray}{$\scriptsize\displaystyle\textcolor{white}{\textbf{NONE}}$}}
        \end{overpic} &  
        \begin{overpic}[width=0.12\textwidth]{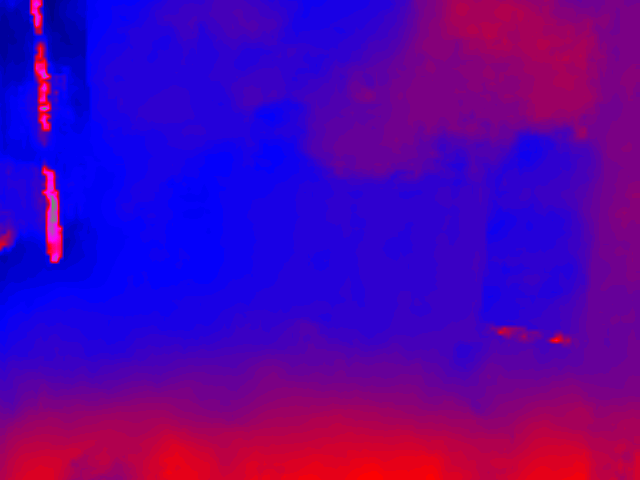}
        \put (0,60) {\colorbox{gray}{$\scriptsize\displaystyle\textcolor{white}{\textbf{MAD++}}$}}
        \end{overpic} &
        \begin{overpic}[width=0.12\textwidth]{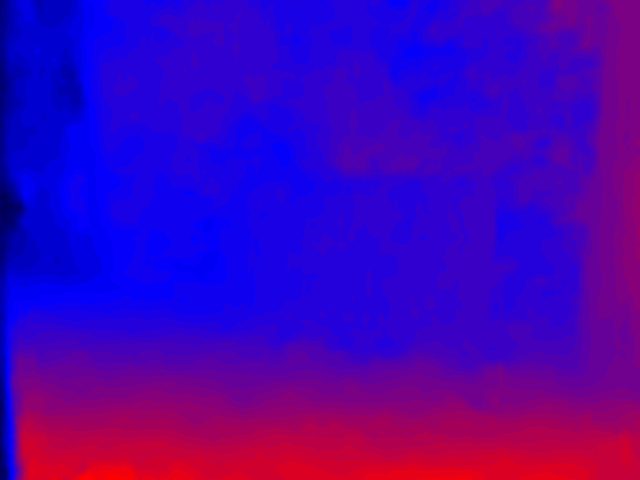}
        \put (0,60) {\colorbox{gray}{$\scriptsize\displaystyle\textcolor{white}{\textbf{K-MAD++}}$}}
        \end{overpic} \\
        \\
        \multirow{2}{*}{\rotatebox{90}{Frame 900}} & \begin{overpic}[width=0.12\textwidth]{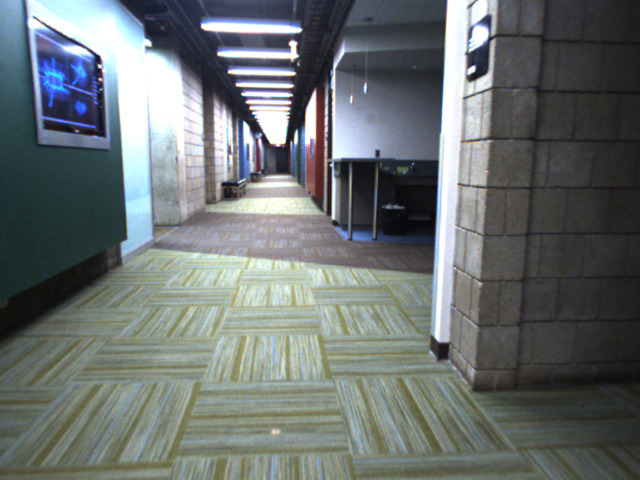} 
        \put (0,60) {\colorbox{gray}{$\scriptsize\displaystyle\textcolor{white}{\textbf{LEFT}}$}}
        \end{overpic} &  
        \begin{overpic}[width=0.12\textwidth]{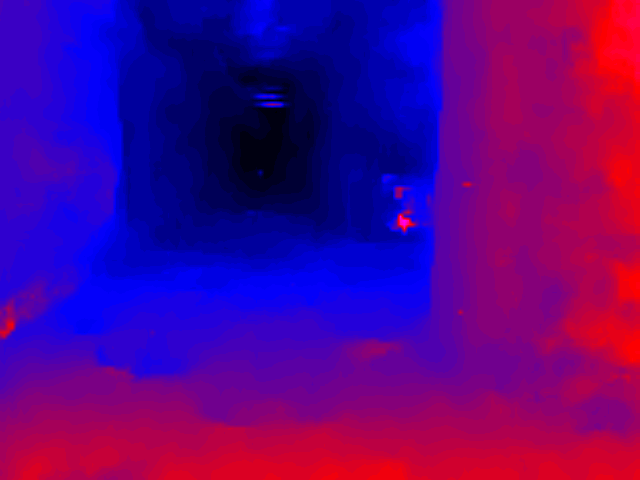}
        \put (0,60) {\colorbox{gray}{$\scriptsize\displaystyle\textcolor{white}{\textbf{MAD}}$}}
        \end{overpic} &
        \begin{overpic}[width=0.12\textwidth]{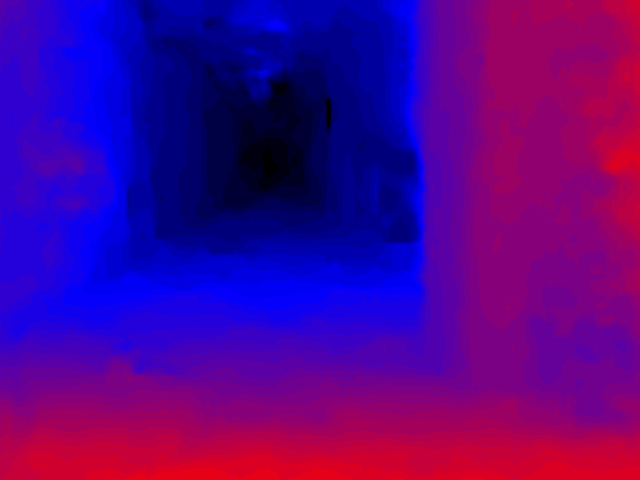}
        \put (0,60) {\colorbox{gray}{$\scriptsize\displaystyle\textcolor{white}{\textbf{K-MAD}}$}}
        \end{overpic} \\
        & \begin{overpic}[width=0.12\textwidth]{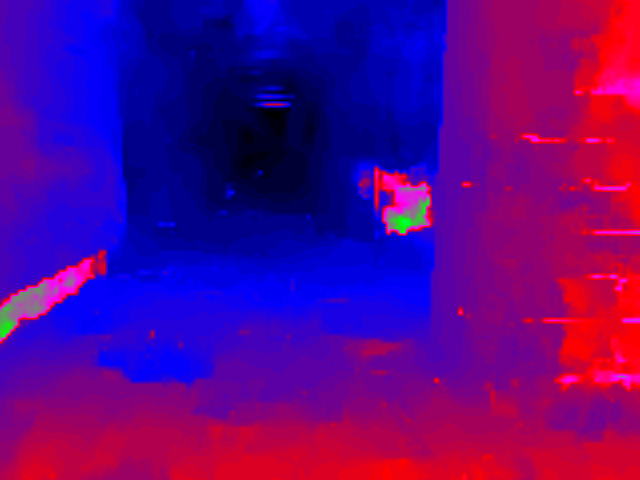} 
        \put (0,60) {\colorbox{gray}{$\scriptsize\displaystyle\textcolor{white}{\textbf{NONE}}$}}
        \end{overpic} &  
        \begin{overpic}[width=0.12\textwidth]{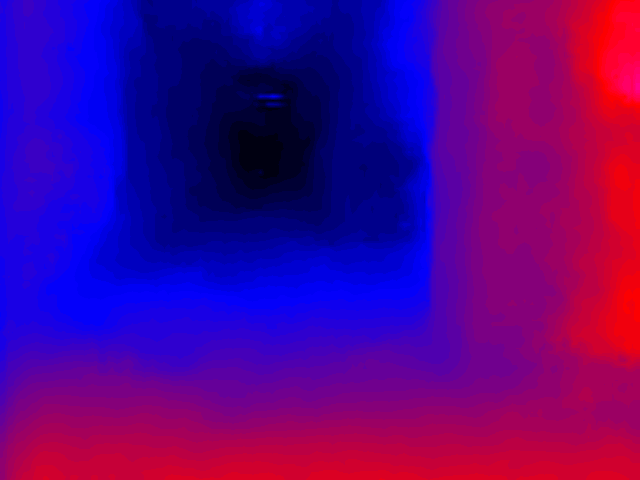}
        \put (0,60) {\colorbox{gray}{$\scriptsize\displaystyle\textcolor{white}{\textbf{MAD++}}$}}
        \end{overpic} &
        \begin{overpic}[width=0.12\textwidth]{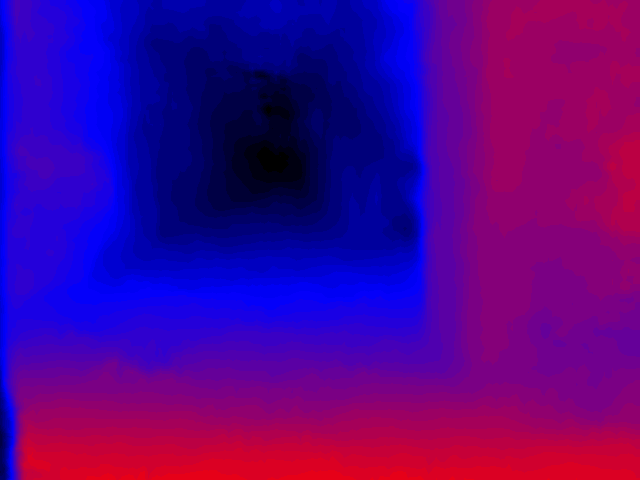}
        \put (0,60) {\colorbox{gray}{$\scriptsize\displaystyle\textcolor{white}{\textbf{K-MAD++}}$}}
        \end{overpic} \\           
    \end{tabular}
    \caption{\rev{\textbf{Qualitative results for different continual adaptation strategies.} We show, across time, the reference image of a stereo pair from WeanHall sequence and different adaptation strategies among those reported in \autoref{tab:weanhall} (MAD++ uses SGM labels).}}
    \label{fig:qualitatives_weanhall}
\end{figure}

\section{Qualitative results}

To conclude, we refer the reader to a video available at \url{https://www.youtube.com/watch?v=YnPGbQE2dLQ} which shows the disparity maps and associated error curves for the methods and datasets considered throughout  the paper.   Starting from KITTI, we point out that, after 20-30 frames of instability due to the domain shift (\ie{}, less than 3 seconds in the video), \algoname{}++ rapidly adapts to the new environment, outperforming \algoname{}. When moving across KITTI sequences, we observe steps  in the error curve upon scene changes, with \algoname{}++ consistently yielding better performance, \ie faster adaptation to the new scene and smaller error. 
When moving to DrivingStereo and WeanHall, we point out how keeping adapting from KITTI (K-\algoname{} and K-\algoname{}++) turns out  more effective than starting the adaptation process from scratch (\algoname{} and \algoname{}++). On DrivingStereo, where the adopted metric is the disparity error with respect to the ground-truth, we observe how  K-\algoname{}++ provides better performance than  K-\algoname{}. \rev{A qualitative example is shown in \autoref{fig:qualitatives_driving}.} On WeanHall, K-\algoname{}++ behaves quite equivalently to  K-\algoname{}, although it is worth pointing out that, due to the lack of ground-truth disparities, the metric adopted to assess performance is exactly the photometric error  minimized by the latter to pursue continual adaptation. \rev{A qualitative example is shown in \autoref{fig:qualitatives_weanhall}.}

\section{Conclusion}

We have presented a novel continual adaptation paradigm for deep stereo networks conceived to deal with challenging and ever-changing environments. By coupling  a \extendednetname{} with a \extendedalgoname{} strategy leveraging on either proxy labels sourced from traditional algorithms or self-supervision via the photometric error, our framework realizes the first-ever real-time and self-adaptive deep stereo network. Experimental results on a variety of datasets support the effectiveness of our proposal, highlighting in particular how proxy supervision is more beneficial than self-supervision and that continual adaptation holds the potential to address the unavoidable  domain shifts that would occur when deploying deep stereo in many practical applications. The experimental findings also provide evidence on the ability of our paradigm to learn knowledge that can transfer well across domains, avoiding, in particular, catastrophic forgetting.

\textbf{Acknowledgments.} We gratefully acknowledge the support of NVIDIA Corporation with the donation of the Titan Xp GPU used for this research.

\ifCLASSOPTIONcaptionsoff
  \newpage
\fi

\bibliographystyle{IEEEtran}
\bibliography{egbib}

\newpage\phantom{Supplementary}
\multido{\i=1+1}{4}{


\section*{Supplementary material}

# Continual Adaptation for Deep Stereo – Supplementary material

Matteo Poggi[1]    Alessio Tonioni[2]    Fabio Tosi[1]
Stefano Mattoccia[1]    Luigi Di Stefano[1]

[1]University of Bologna, Italy    [2]Google Inc.

In this document, we provide additional experiments concerning "Continual Adaptation for Deep Stereo", complementing those reported in the main paper.

## 1. Batched adaptation

As discussed in the main paper, updating the network using batches of images instead of individual ones might improve performance. However, this usually happens because building a batch allows to optimize the network over a set of variegate samples, that should better approximate the entire training set and thus lead to more stable gradients compared to those computed on a single sample. In our setting, however, the update steps occur on images that are acquired closely in terms of time and are, therefore, very similar. Thus, the gradients provided by the individual samples are quite similar the average gradient that would be computed in a batch. We show an experiment to investigate on whether updating the network based on a batch of 10 samples, which slows down the inference rate by a factor of about 4, may bring in significant improvements. The results are shown in Table 1: by comparing the first and second column pairs, one can observe how batch-updating (second column pair) provides just a very small performance improvement wrt our approach (first pair). On the other hand, one might recover full-speed by updating the network every 10 frames based on a batch of 10 samples, but this deteriorates accuracy significantly, as shown in the third column pair of Table 1.

| | | step=1 batch=1 | | step=1 batch=10 | | step=10 batch=10 | |
|---|---|---|---|---|---|---|---|
| Adapt. Mode | Proxy src. | D1 | EPE | D1 | EPE | D1 | EPE |
| FULL | ✗ | 2.43 | 0.95 | 2.35 | 0.95 | 4.43 | 1.21 |
| FULL++ | SGM | 2.28 | 0.95 | 2.25 | 0.95 | 4.44 | 1.18 |
| *MAD* | ✗ | 4.09 | 1.19 | 4.03 | 1.18 | 11.37 | 2.31 |
| *MAD++* | SGM | 2.46 | 0.98 | 2.43 | 0.97 | 7.36 | 1.41 |

Table 1. **Adaptation on image batches.** Experiment dealing with continual adaptation from synthetic pretraining on *Campus* → *City* → *Residential* → *Road*. We perform updates with single images (batch=1) or stacking the latest 10 (batch=10), at any time a new sample is available (step=1) or after collecting 10 new samples (step=10). Pretraining on synthetic images.

1

## 2. Sensitivity of parameters $\delta$ and $\lambda$ of Algorithm 2

The two parameters $\delta$ and $\lambda$, respectively a decay and an update weight, are introduced in Section 3.1 to gradually decay the probability distribution over time and balance the impact of a successful update, respectively. We have tuned them to 0.99 and 0.01 after an analysis carried out on Campus (the shortest and hardest sequence) and starting from *MADNet* weights trained on synthetic data. Table 2 shows D1 error for different decay-weight configurations, highlighting better results with 0.99, 0.01. However, as also noticeable in Table 2, we found that these hyper-parameters are not critical for making adaptation work.

|  |  | \multicolumn{4}{c}{decay} |  |  | \multicolumn{4}{c}{decay} |
|---|---|---|---|---|---|---|---|---|---|---|---|
|  |  | 1 | 0.99 | 0.75 | 0.50 |  |  | 1 | 0.99 | 0.75 | 0.50 |
| weight | 1 | 22.86 | 22.35 | 22.59 | 22.72 | weight | 1 | 12.35 | 12.78 | 12.70 | 14.31 |
|  | 0.1 | 22.59 | 22.31 | 22.55 | 22.62 |  | 0.1 | 11.70 | 12.69 | 13.35 | 11.53 |
|  | 0.01 | 22.35 | **22.27** | 22.37 | 22.36 |  | 0.01 | 13.65 | **11.50** | 13.24 | 13.00 |
|  | 0.001 | 22.40 | 22.40 | 22.46 | 22.68 |  | 0.001 | 12.38 | 12.02 | 11.95 | 11.80 |
|  |  | \multicolumn{4}{c}{MAD} |  |  | \multicolumn{4}{c}{MAD++} |

Table 2. **Hyper-parameters tuning** on Campus sequence (1149×2 frames), with *MADNet* pretrained on synthetic data.

## 3. Proxy supervision vs meta-learning.

Following the L2A paper, we take a *MADNet* model trained on FlyingThings3D and 1) fine-tune it on Synthia in a traditional, supervised manner or 2) using the First Order approximation variant of L2A (FOL2A). We tried also to use the full meta learning formulations without approximation (L2A and L2A-Wad), but unfortunately, the training of the network was very unstable and always lead to diverging models. Thus, for this experiment, 1) provides the baseline with respect to which we evaluate the adaptation performance provided by FOL2A and our proposed approach dealing with proxy labels gathered by SGM. Table 3 collects the outcome of this experiment on the KITTI raw dataset. We can notice how using L2A during pre-training can improve the performance with respect to the baseline. However, our proposal turns out significantly more effective, *i.e.* proxy supervision is more effective than L2A pre-training. For the sake of space, we report in the main paper only the numbers obtained over the whole dataset (All entry in Table 3), while Table 3

| Starting Model | Adapt. Mode | City (8027 frames) | | Residential (28067 frames) | | Campus (1149×2 frames) | | Road (5674 frames) | | All | |
|---|---|---|---|---|---|---|---|---|---|---|---|
|  |  | D1-all(%) | EPE | D1-all(%) | EPE | D1-all(%) | EPE | D1-all(%) | EPE | D1-all(%) | EPE |
| *MADNet*-Synthia | FULL (baseline) | 3.99 | 1.26 | 3.14 | 1.10 | 15.09 | 2.55 | 3.89 | 1.19 | 2.76 | 1.03 |
| *MADNet*-Synthia-FOL2A [1] | FULL | 3.90 | 1.25 | 2.93 | 1.06 | 13.98 | 2.49 | 3.76 | 1.21 | 2.60 | 1.01 |
| *MADNet*-Synthia | FULL++ (ours) | 3.58 | 1.14 | 2.29 | 0.95 | 10.62 | 1.79 | 3.04 | 1.05 | 2.23 | 0.95 |

Table 3. **Comparison of online adaptation strategies across different domains**. Results on the *City*, *Residential*, *Campus* and *Road* sequences from KITTI. Comparison between *MADNet* trained on FlyingThings3D and finetuned on Synthia, with (FOL2A) or without meta-learning optimization.

# 4. Complementing low-density supervision.

We have carried out additional experiments by combining WILD labels with LIDAR as well with self-supervision, so as to study whether one might effectively complement the low density supervision provided by WILD. As for the former setting, we use LIDAR measurements where WILD proxy labels are missing due to low confidence. This allows to consider LIDAR points only where WILD does not provide labels (e.g., untextured regions), while using WILD where LIDAR usually fail, *i.e.* close to object boundaries. In the latter, we just add the two losses, as also proposed in previous works [2, 3]. Results are collected in Table 4. We can notice that complementing WILD with LIDAR supervision consistently improves accuracy. As for self-supervision, this tend to be the case mostly for FULL, while MAD seems to benefit when adapted by the stronger supervision provided by a regression loss, based on either proxy labels, LIDAR measurements or both.

| Adapt. Mode | Source | City (8027 frames) D1-all(%) | EPE | Residential (28067 frames) D1-all(%) | EPE | Campus (1149×2 frames) D1-all(%) | EPE | Road (5674 frames) D1-all(%) | EPE | All D1-all(%) | EPE |
|---|---|---|---|---|---|---|---|---|---|---|---|
| FULL++ | WILD+Self-Sup. | 3.08 | 1.06 | 2.13 | 0.91 | 8.04 | 1.49 | 2.34 | 0.96 | 2.04 | 0.92 |
| FULL++ | WILD+LIDAR | 3.58 | 1.02 | 2.37 | 0.91 | 9.09 | 1.47 | 3.17 | 0.92 | 2.18 | 0.89 |
| MAD++ | WILD+Self-Sup. | 5.56 | 1.29 | 4.30 | 1.13 | 12.13 | 1.87 | 5.16 | 1.20 | 3.09 | 1.00 |
| MAD++ | WILD+LIDAR | 4.10 | 1.08 | 2.66 | 0.93 | 11.28 | 1.76 | 4.71 | 1.15 | 2.43 | 0.91 |

Pretraining on **synthetic** data

| Adapt. Mode | Source | City D1-all(%) | EPE | Residential D1-all(%) | EPE | Campus D1-all(%) | EPE | Road D1-all(%) | EPE | All D1-all(%) | EPE |
|---|---|---|---|---|---|---|---|---|---|---|---|
| FULL++ | WILD+Self-Sup. | 1.51 | 0.88 | 1.75 | 0.85 | 3.80 | 1.11 | 1.08 | 0.80 | 1.66 | 0.85 |
| FULL++ | WILD+LIDAR | 1.48 | 0.79 | 1.28 | 0.76 | 3.26 | 0.95 | 1.19 | 0.75 | 1.25 | 0.75 |
| MAD++ | WILD+Self-Sup. | 1.65 | 0.91 | 2.97 | 0.95 | 4.22 | 1.25 | 1.34 | 0.85 | 2.20 | 0.91 |
| MAD++ | WILD+LIDAR | 1.43 | 0.80 | 1.67 | 0.80 | 3.68 | 1.10 | 1.19 | 0.76 | 1.34 | 0.77 |

Pretraining on **synthetic** data, fine-tuning on KITTI 2015 GT.

Table 4. **Online adaptation within a single domain (WILD+Lidar labels)**. Results on the *City*, *Residential*, *Campus* and *Road* sequences from KITTI.

# 5. Comparison with sparse (and filled) SGM

As we have observed in the experiments on the DrivingStereo sequences, the sparse proxy labels generated by SGM and WILD are very accurate, although they have low density. However, we point out that lower density might be rather undesirable in real applications. For instance, let us consider an autonomous driving scenario, as depicted in Figure 1 below from the Rainy sequence used in our experiments. Although selecting the proxy algorithm with higher density, the sparse SGM map lacks measurements in critical areas close to the vehicle, whilst the smooth and dense map perceived by our method seems quite more amenable to support high-level tasks dealing with planning and navigation.

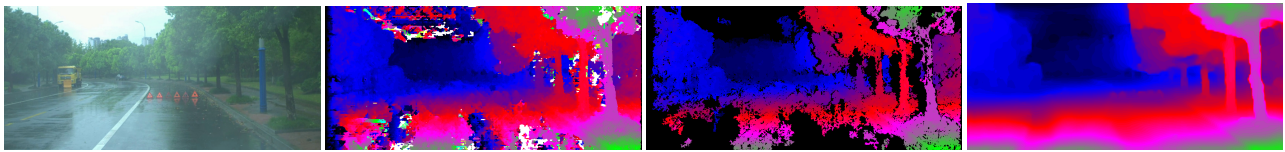

Reference image      SGM (before filtering)      SGM (after filtering)      *MADNet*-K (MAD++)

Figure 1. **Qualitative results achieved by SGM and *MADNet*.** Example taken from *Rainy* sequence.

Given the good accuracy provided by SGM sparse proxies, one might try to densify these sparse depth measurements in order to use them in practical applications. Thus. we have performed depth completion on the sparse SGM maps by a standard hand-crafted interpolation method (i.e., the hole-filling algorithm used to densify disparity maps when submitting to the KITTI online banchmark) and a recent learning-based solution (the network used in [4]) trained on the same synthetic data as *MADNet*. We report the results of this experiment in Table 5. We notice that hole-filling can outperform our proposal in the Cloudy sequence only, i.e. the one where SGM achieves the lowest error among the three sequences, as shown in the main paper. However, more challenging conditions (as observed in the Rainy and Dusky sequences) are better tackled by *MADNet* thanks to online adaptation. We would also like to highlight, that, as shown in Figure 2, standard hand-crafted hole filling is particularly prone to yield gross artifacts that do not affect the error metrics due to the sparsity of the ground-truth provided with DrivingStereo (filtered LIDAR). Yet, more advanced learning-based depth densification methods seem significantly affected by the domain shift, as vouched by the second row in Table 5.

| Completion method | Rainy (1667 frames) D1-all(%) | EPE | Dusky (1119 frames) D1-all(%) | EPE | Cloudy (4950 frames) D1-all(%) | EPE |
|---|---|---|---|---|---|---|
| SGM + Hole-filling† | 12.68 | 3.05 | 7.88 | 1.30 | 4.50 | 1.19 |
| SGM + ECCV 2020 | 17.90 | 3.91 | 14.53 | 1.91 | 9.74 | 1.83 |
| *MADNet*-K-MAD++ (SGM) | 12.65 | 2.32 | 5.93 | 1.40 | 6.26 | 1.72 |

Table 5. **Comparison between online adaptation and SGM densification on DrivingStereo.** Results on the *Rainy*, *Dusky* and *Cloudy* sequences. Learned methods are trained on synthetic data. †means non-learned algorithm.

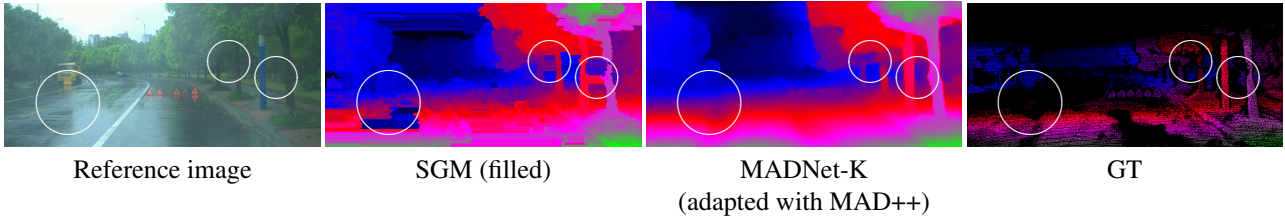

Reference image | SGM (filled) | MADNet-K (adapted with MAD++) | GT

Figure 2. **Comparison between SGM densification and online adaptation.** Example taken from *Rainy* sequence.


}

\end{document}